\documentclass{article}
\usepackage{times}
\usepackage{graphicx}
\usepackage{subfigure}
\usepackage[numbers]{natbib}
\usepackage{algorithm}
\usepackage{algorithmic}
\usepackage{amsmath}
\usepackage{amsthm}
\usepackage{amsfonts}
\usepackage{amssymb}
\usepackage{multirow}

\def\real{\mathbb{R}}
\def\D{\mathbb{D}}
\def\S{\mathbb{S}}
\def\E{\mathbb{E}}
\def\P{\mathbb{P}}
\def\N{\mathcal{N}}
\def\argmax{\mathop{\mathrm{argmax}}}
\def\argmin{\mathop{\mathrm{argmin}}}

\newtheorem{theorem}{Theorem}[section]
\newtheorem{lemma}[theorem]{Lemma}

\title{A Lipschitz Exploration-Exploitation Scheme for Bayesian Optimization} 

\author{ Ali Jalali \\  
Turn Inc \\  
ajalali@turn.com\\ 
$\qquad\qquad\qquad\qquad\qquad\qquad$\and 
Javad Azimi \\
Microsoft Inc\\
jaazimi@microsoft.com\\ 
$\qquad\qquad\qquad\qquad\qquad\qquad$\and
Xiaoli Fern \\ 
Oregon State University\\ 
xfern@eecs.oregon.edu\\
$\qquad\qquad\qquad\qquad\qquad\qquad$\and
Ruofei Zhang \\
Microsoft Inc\\
bzhang@microsoft.com\\ 
} 
 
\begin{document} 
 
\maketitle 
 
\begin{abstract}
The problem of optimizing unknown costly-to-evaluate functions has been studied extensively in the context of Bayesian optimization. Algorithms in this field aim to find the optimizer of the function by requesting only a few function evaluations at carefully selected locations. An ideal algorithm should maintain a perfect balance between exploration (probing unexplored areas) and exploitation (focusing on promising areas) within the given evaluation budget. In this paper, we assume the unknown function is Lipschitz continuous. Leveraging the Lipschitz property, we propose an algorithm with a distinct exploration phase followed by an exploitation phase. The exploration phase aims to select samples that shrink the search space as much as possible, while the exploitation phase focuses on the reduced search space and selects samples closest to the optimizer. We empirically show that the proposed algorithm significantly outperforms the baseline algorithms.
\end{abstract}

\section{Introduction}

In many applications, we would like to optimize an unknown function $f(\cdot)$ that is costly to evaluate over a compact input space. Classic optimization methods, such as gradient descent, cannot be applied to this type of problems since they need to evaluate the function frequently. In contrast, Bayesian Optimization (BO) \cite{taxonomy,Bayesian09} algorithms try to solve this problem with a small number of function evaluations. Bayesian optimization algorithms, generally, have two key components: 1) A posterior model to predict the output value of the function at any arbitrary input point, and 2) A selection criterion to determine which point to be evaluated next. 

The first step of a BO algorithm is to learn a posterior probabilistic model over unobserved points of the function. Gaussian processes (GP) \cite{Rasmussen06} have been used in the literature of Bayesian optimization as the probabilistic posterior model. GP models the function output for any unobserved point in the input space as a normal random variable, whose mean and variance depend on the location of the point in relation to a set of given observed samples. Based on the learned posterior model, a selection criterion is then used to choose the next sample to be evaluated. A number of selection criteria have been proposed in the literature of Bayesian optimization. They typically work by selecting an example that optimizes some objective function designed to balance between exploring unobserved area and exploiting areas that are promising based on existing observations. Maximum probability of improvement \cite{Elder92,Stuckman88} and maximum expected improvement (EI) \cite{Locatelli97} are two successful examples.

In this paper, we focus on the design of the selection criterion for Bayesian optimization. In particular, we study BO in a sequential setting~\cite{taxonomy,Moore98}, where the samples are chosen sequentially and a selection is made only after the function evaluations of the previous samples are revealed. 
We make a mild assumption that the unknown function is Lipschitz-continuous. Leveraging the Lipschitz property, we design a selection algorithm that operates in two distinct phases: the exploration phase and the exploitation phase. In general, in the context of Bayesian optimization \cite{taxonomy} and bandit problems \cite{Bruce10}, the exploration phase selects sample from unexplored area while the exploitation focuses on promising area. In this paper, we introduce a new interpretation of exploration and exploitation.

The exploration phase of the proposed algorithm, at each step, selects a sample that eliminates the largest possible portion of the input space while guaranteeing, with high probability, that the eliminated part does not include the maximizer of the function. Hence, the exploration stage of the algorithm tries to shrink the search space of the function as much as possible. In contrast, the exploitation phase of our algorithm selects the point which is believed to be the closest sample to the optimal point with high probability. 

Experimental results over $8$ real and synthetic benchmarks indicate that the proposed approach is able to outperform the Expected Improvement (EI) criterion, one of the current state-of-the-art BO selection methods. In particular, we show that our algorithm is better than EI both in terms of the mean and variance of the performance. We also investigate whether combining our exploration stage with EI can boost the performance of EI. However, the results were negative. Sometimes it helps and sometimes it hurts and on average we observe little to no improvement to EI. This is possibly because our exploration method actively aims to eliminate regions from the input space and the EI criterion does not take that into consideration when selecting samples.



The remainder of the paper is organized as follows. In Section \ref{sec:motivation}, we motivate the use of exploration-exploitation Bayesian optimization by analyzing the behavior of EI. Section \ref{sec:FHBO} introduces our algorithm and provides insights into both theoretical and practical aspects of the algorithm. Experimental evaluation of our algorithm is shown in Section \ref{sec:results}. Finally, the paper is concluded in Section \ref{sec:conclusion}.

\section{Motivating Observation}
\label{sec:motivation}
\def\O{\mathcal{O}}

In this section, we motivate our approach by revealing a key observation about the well known Expected Improvement (EI) algorithm \cite{Locatelli97}. The original EI is defined as 
\begin{equation}
EI(x) =  \mathbb{E} \left[(f(x)-y_{\max})\, \mathbb{I}_{\{f(x)-y_{\max}>0\}}\right],
\end{equation}
where $\mathbb{I}_{\{\cdot\}}$ is the indicator function. Hence, it measures the expected improvement of the choice of $x$ over the current maximum function evaluations $y_{\max}$ over observed samples.Using Gaussian Process (GP) \cite{Rasmussen06} as the posterior model of the unknown function, the EI objective can be represented by
\begin{equation}
\begin{aligned}
EI(x|\O) &= (\mu_{x|\O} - y_{\max})\Phi\left(\frac{\mu_{x|\O} - y_{\max}}{\sigma_{x|\O}}\right)+\sigma_{x|\O}\,\phi\left(\frac{\mu_{x|\O} - y_{\max}}{\sigma_{x|\O}}\right),
\end{aligned}
\label{eq:EI-def}
\end{equation}
where, $\mu_{x|\O}$ and $\sigma_{x|\O}$ are the mean and standard deviation associated with the point $x$ by GP, and, $\Phi(\cdot)$ and $\phi(\cdot)$ are standard Gaussian CDF and PDF, respectively. Here, $\O=\{(x_i,f(x_i))\}_{i=1}^n$ is the set of $n$ observed samples $x_{\O}$ with their function evaluations $f(x_{\O})$ and define $y_{\max}=\max_{x_i\in x_{\O}} f(x_i)$. Further, the means and variances are defined as follows:
\begin{equation}
\begin{aligned}
\mu_{x|\O} &= k(x,x_{\O})\,k(x_{\O},x_{\O})^{\!-1}\,f(x_{\O})\\
\sigma^2_{x|\O} &= k(x,x) - k(x,x_{\O})\,k(x_{\O},x_{\O})^{\!-1}\,k(x_{\O},x),
\end{aligned}
\nonumber
\end{equation}
where $k(\cdot,\cdot)$ is some kernel function. In this paper, we consider Gaussian kernel $k(x_1,x_2) = \exp(-\frac{1}{\ell}\|x_1-x_2\|_2^2)$.\\

EI has been widely used and studied; however, there has been always a concern about balancing the exploration and exploitation of EI. The main reason for this concern is that even though the asymptotic convergence of EI is guaranteed under certain conditions \cite{Vasquez10}, EI tries to exploit the information and potentially can request a lot of samples if it hits a local optimum region, while we have a limited number of experiments. There has been some attempts in the literature to address this concern with varying degrees of success, which we briefly discuss here.

\begin{itemize}
\label{eq:ei}
\item [(a)]  Considering the original definition of EI, researchers have proposed to replace $y_{\max}$ with a smaller value to make EI more exploitative and with a larger value to make it more explorative. In particular, \cite{Liz08} suggested $y_{\max} + \xi$ and \cite{azimi10} suggested $(1+\xi) y_{\max}$ to replace $y_{\max}$. However, this approach has not seen much empirical success. \cite{Liz08} showed that starting with large values of $\xi$ (to be explorative in the beginning) and cooling it down (to make it more and more exploitative) makes little or no difference in the performance of EI.\\

\item [(b)] On a separate line of work, \cite{Schon97} proposed to consider a surrogate function $$EI_\xi(x) = \mathbb{E}\left[(f(x)-y_{\max})^\xi\,\mathbb{I}_{\{f(x)-y_{\max}>0\}}\right].$$
For $\xi=1$, this objective tries to improve over $y_{\max}$ (exploiting mode) and if we decrease $\xi$ it starts to explore uncertain areas (exploration mode). This method is very sensitive to small changes in $\xi$ and except for very specific setup like the one used in \cite{Sasena02}, there is no systematic way to choose $\xi$. This makes it nearly impossible to use this method.

\item [(c)] The third proposal is to have a ``random" exploration phase proceeding EI. In this approach, we take a number of random samples before switching to EI. We analyzed this method in Fig.~\ref{fig:1}. For a fixed budget $n_b$, we run $n_b$ experiments as follows: first we consider the case where there is $1$ random sample followed by $n_b-1$ samples selected by the EI criterion, next we consider the case where there are $2$ random samples followed by $n_b-2$ EI samples and so on.  The purpose of this investigation is to understand whether exploring with random samples prior to selecting with EI can improve the performance of EI, and if so how much exploring is necessary. We run this experiments on a number of different functions introduced in Section~\ref{sec:exper}. These experiments reveal that ``random" exploration never helps EI, since the regret monotonically increases as we increase the number of random samples from 1 to $n_b$. One possible explanation for this behavior is that the values of the function are highly correlated and hence, uniform sampling does not efficiently represent the skewness of the data points.

\begin{figure*}[t]
\begin{center}
\makebox[\textwidth][c]{
\addtolength{\tabcolsep}{-0.7em}
\begin{tabular}{c c c c}
\includegraphics[width=1.7in, height=1.45in]{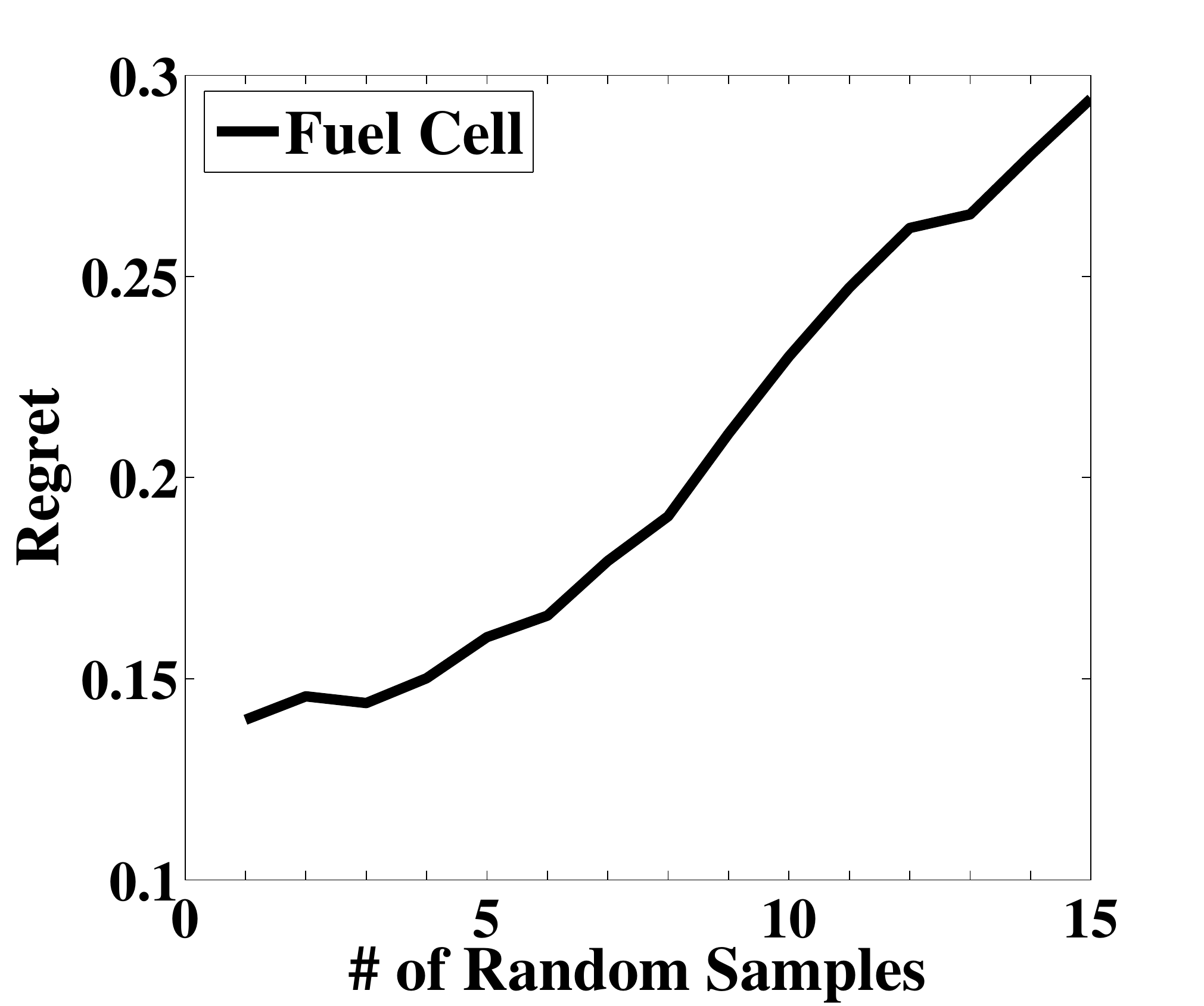}&
\includegraphics[width=1.7in, height=1.45in]{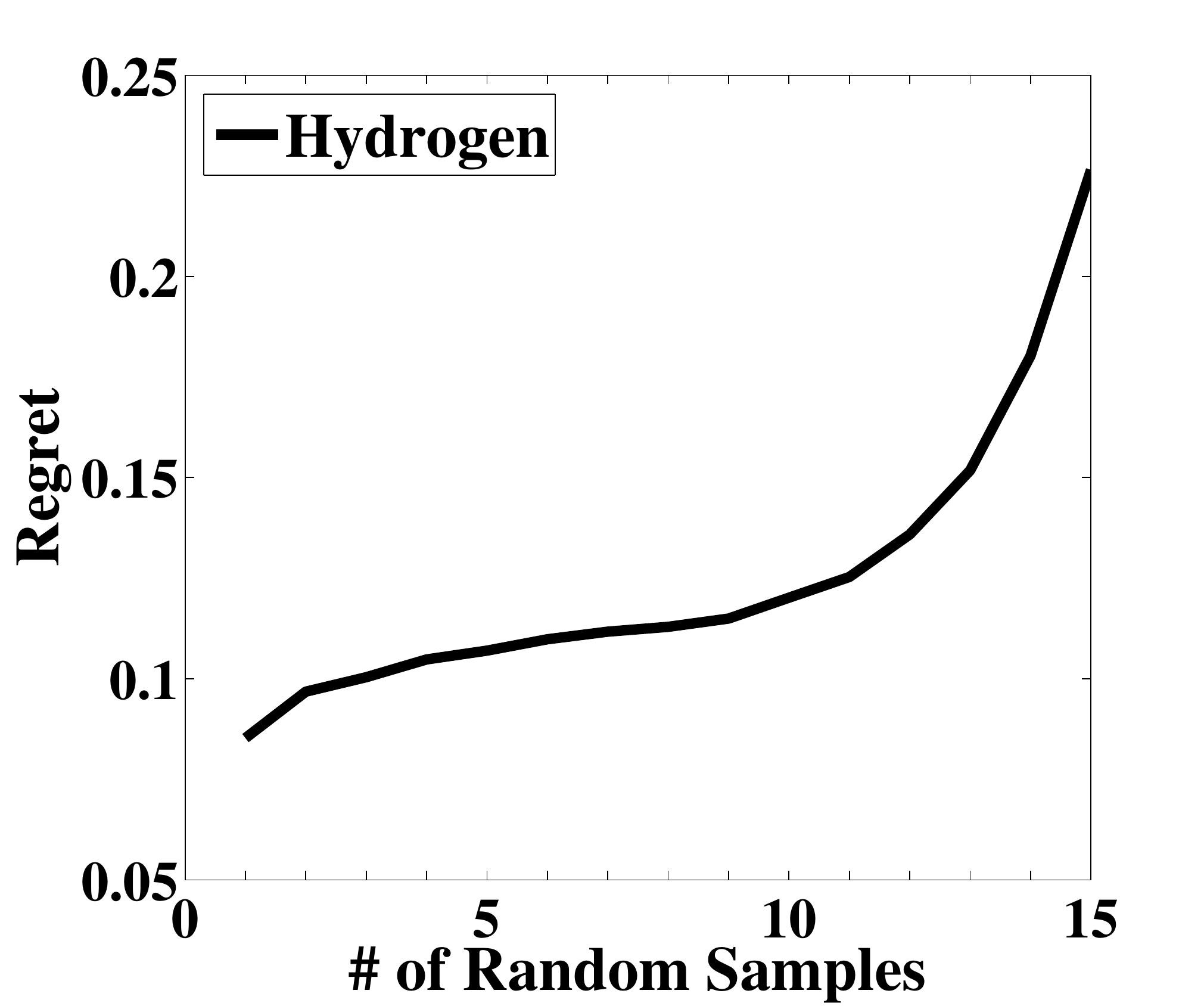}&
\includegraphics[width=1.7in, height=1.45in]{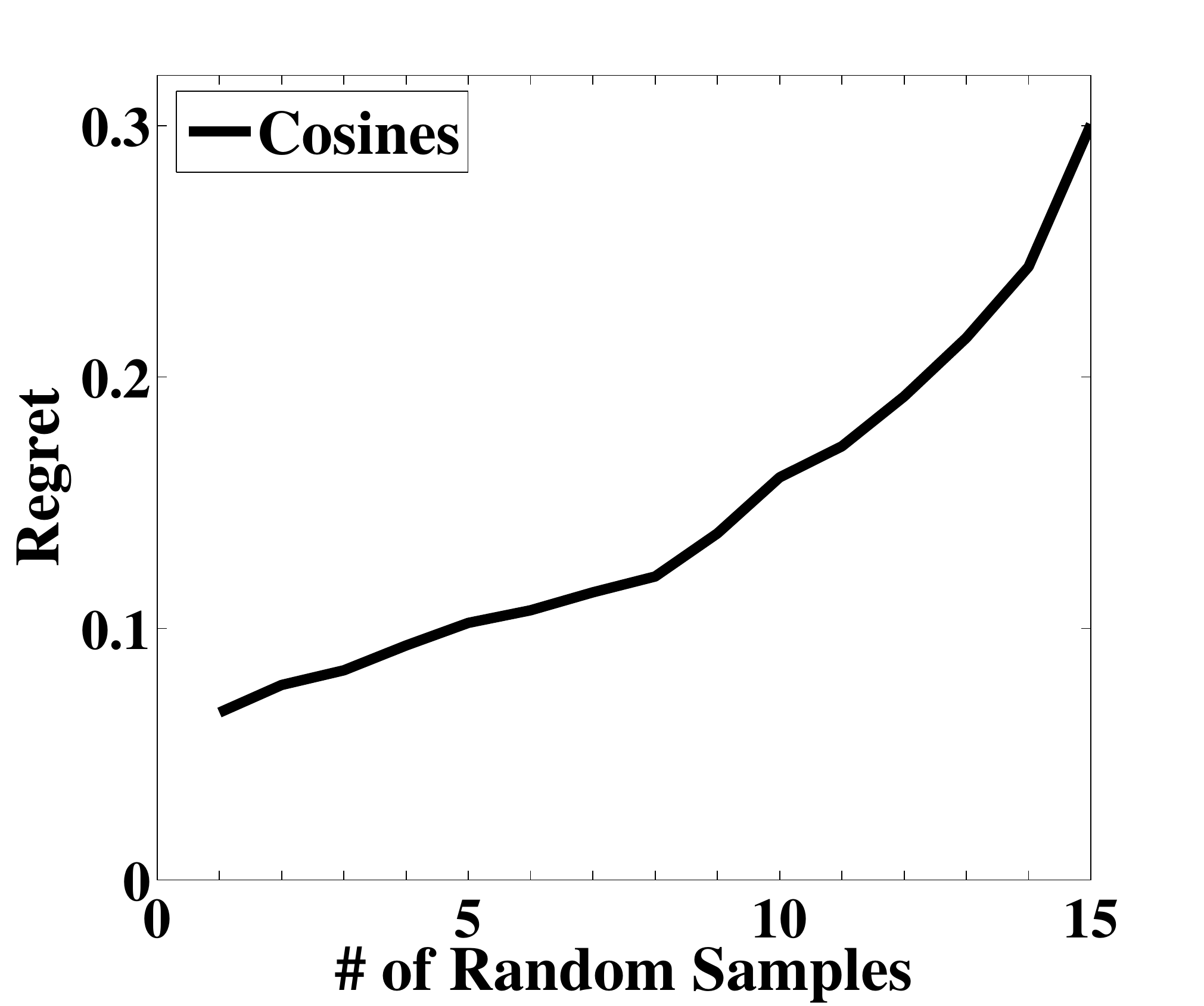}&
\includegraphics[width=1.7in, height=1.45in]{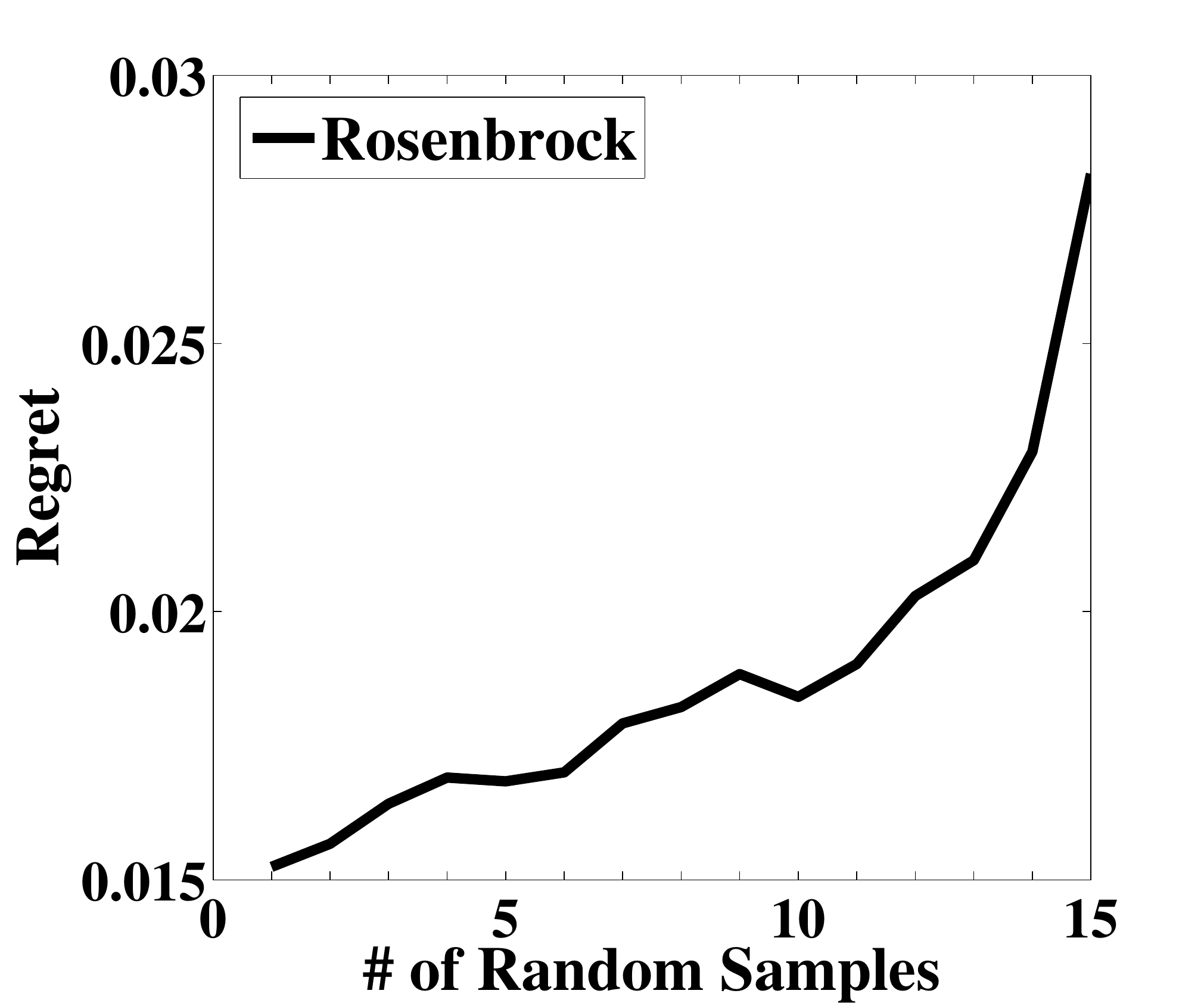}\\
\includegraphics[width=1.7in, height=1.45in]{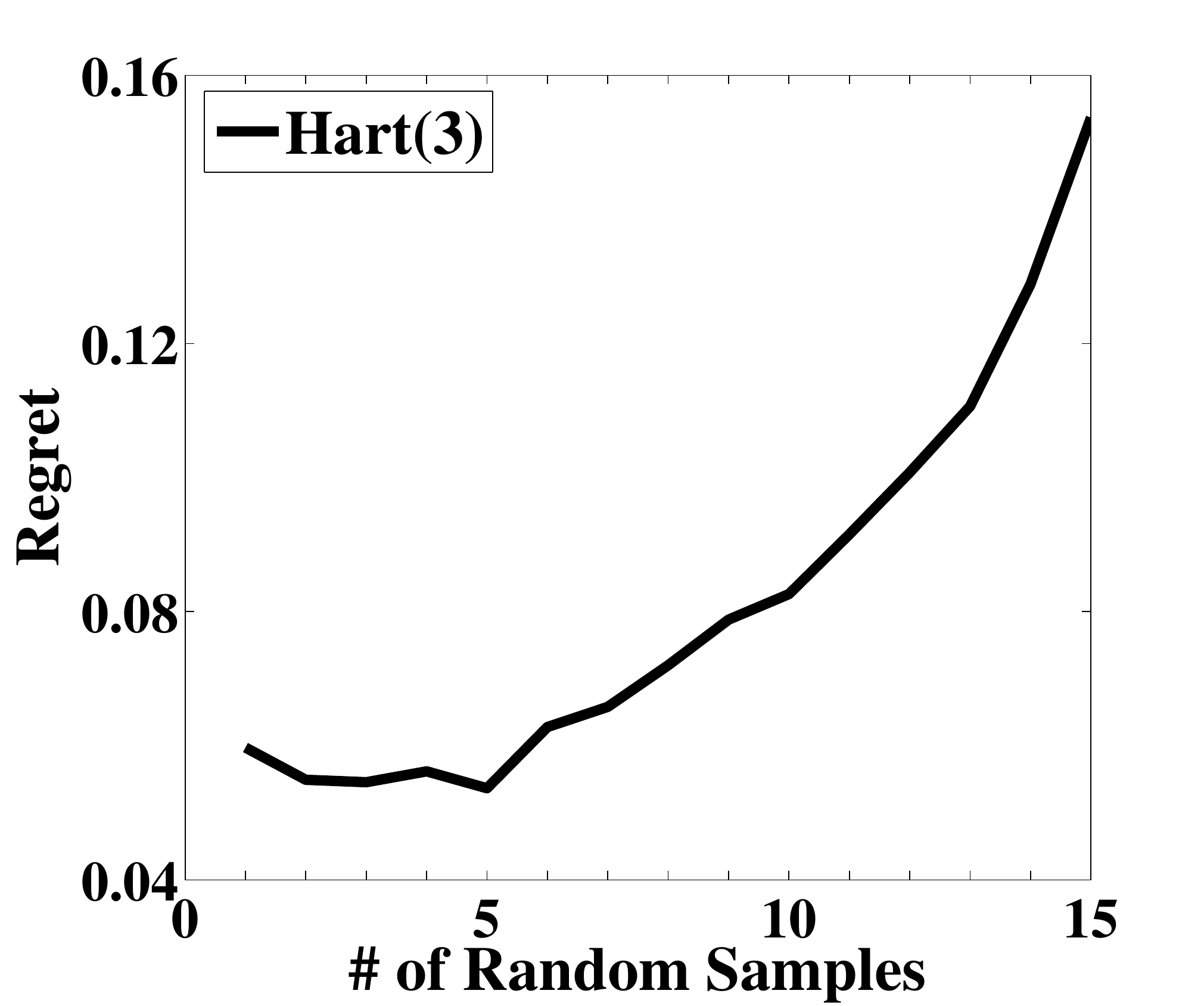}&
\includegraphics[width=1.7in, height=1.45in]{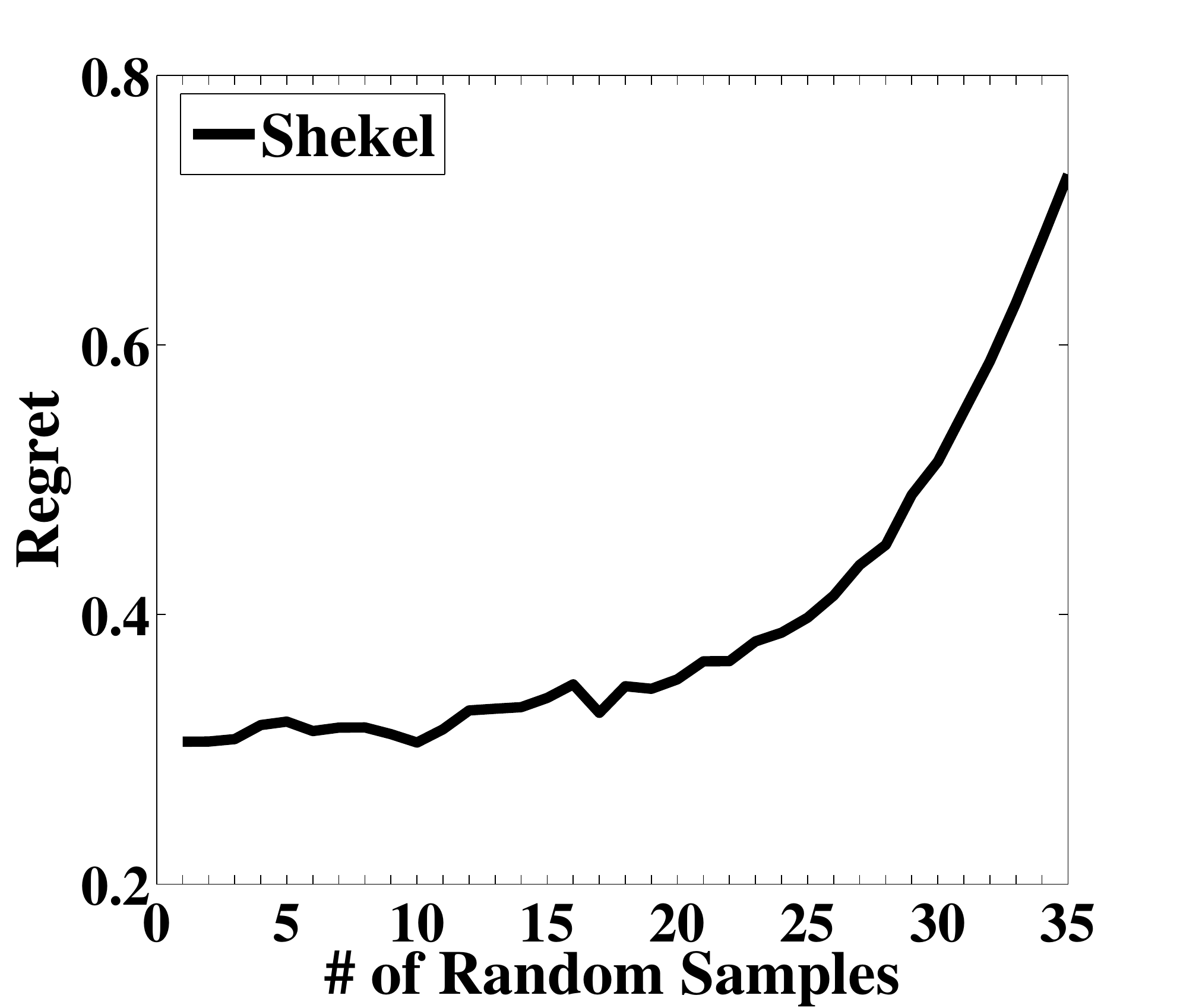}&
\includegraphics[width=1.7in, height=1.45in]{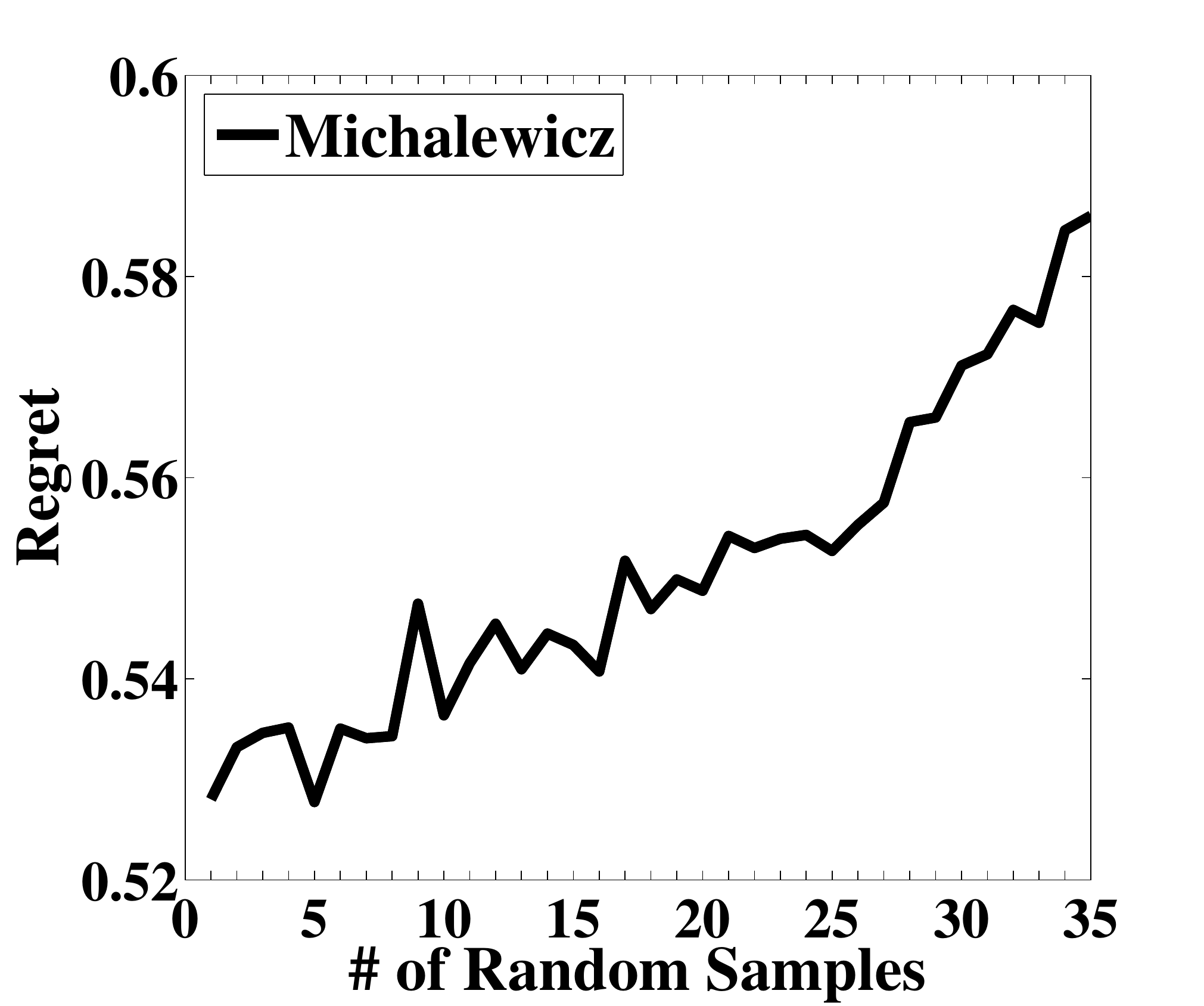}&
\includegraphics[width=1.7in, height=1.45in]{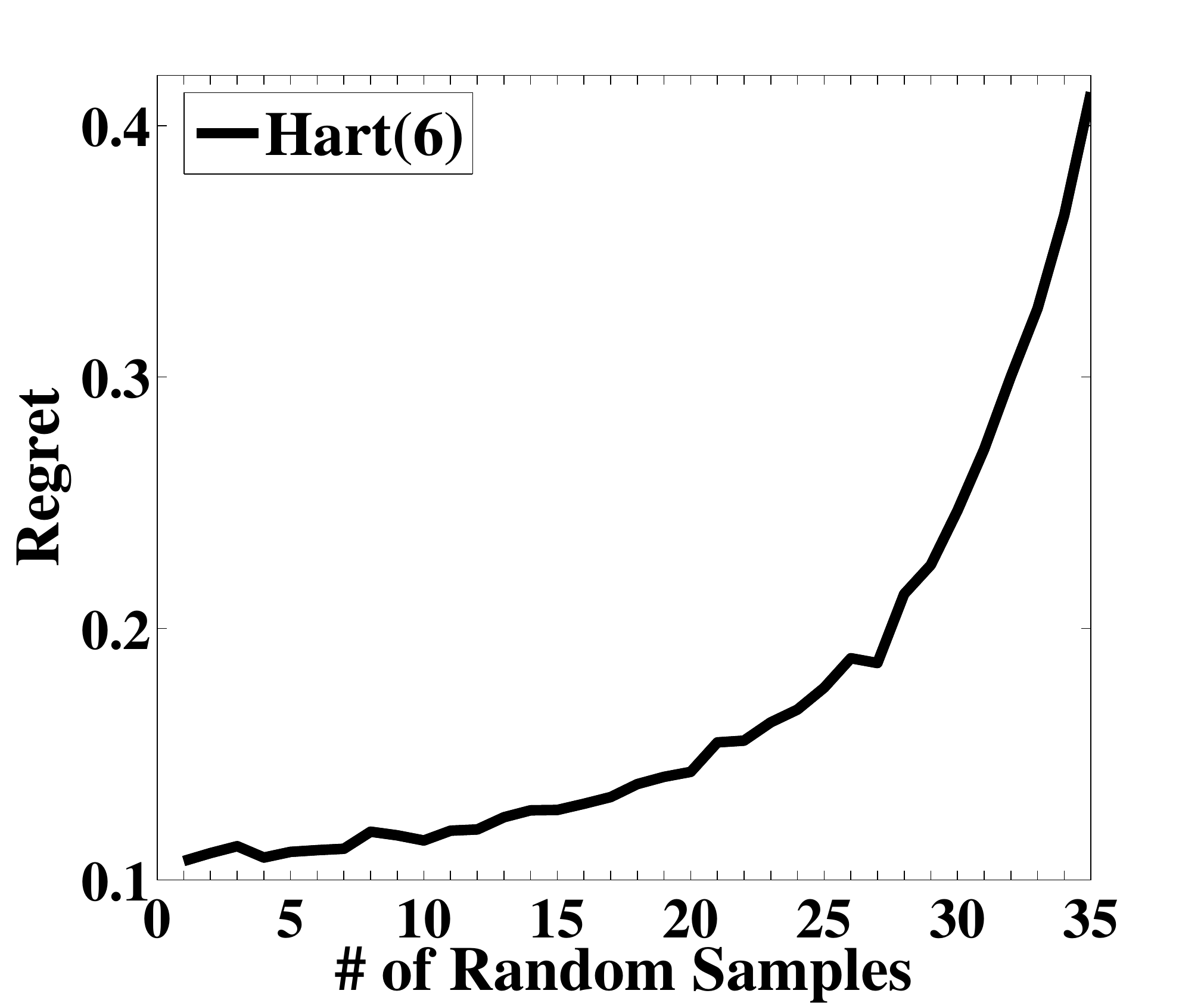}\\
\end{tabular}}
\end{center}
\caption{Plot of regret versus the number of random exploration for EI algorithm. For a fixed budget $n_b$, we run a number of experiments as follows: first we consider the case where there are $1$ random samples followed by $n_b-1$ EI samples, next we consider the case where there are $2$ random samples followed by $n_b-2$ EI samples and so on. For 2D and 3D functions, we let $n_b=15$ and for high-dimensional functions, we let $n_b=35$. This result shows that the best EI performance is when we do not do random exploration.}
\label{fig:1}
\end{figure*}


\end{itemize}

Based on the existing literature as well as our empirical investigation of EI discussed above, we would like to know whether or not it is possible to design an algorithm that operates in two naturally defined phases of exploration and exploitation and achieves consistently better performance than EI. We devote the next section to answer this question and introduce our proposed algorithm.

\section{Finite Horizon Bayesian Optimization}
\label{sec:FHBO}
Not being able to balance the exploration-exploitation, EI might have poor performance especially when the query budget is small. In this section, we propose a two-phase exploration/exploitation algorithm that outperforms EI with its smart exploration and exploitation.

\subsection{Exploration}

\begin{algorithm}[t]

\caption{Next Best exploRative Sample (NBRS)}
\label{alg:explore}
\begin{algorithmic}
\STATE {\bf Input}: Maximum $M$, Lipschitz Constant $L$ and Set of observed samples $\{(x_1,f(x_1)),\ldots,(x_t,f(x_t))\}$
\STATE {\bf Output}: Next best explorative sample $x$
\STATE
\STATE $\displaystyle\D_t = \D - \bigcup_{i=1}^t\S(x_i,r_{x_i})$
\STATE
\STATE \small$\displaystyle x\longleftarrow\argmax_{x\in\D_t}\; \textbf{Vol}\left(\D_t\!\cap\!\S\left(x,\frac{\left|M-\mu_{x|\O}\right|-1.5\sigma_{x|\O}}{L}\right)\right)$\normalsize
\STATE
\STATE
\end{algorithmic}
\end{algorithm}

Generally, a good exploration algorithm should be able to shrink the search space, so that we are left with a small region to focus on during the exploit stage. Let $\D=\bigotimes[a_i,b_i]\in\real^d$ be the Cartesian product of intervals $[a_i,b_i]$ for some $a_i<b_i$ and $i\in\{1,2,\ldots,d\}$. Suppose the unknown function $f:\D\mapsto[m,M]$ (with $f(x^*)=M$) is a Lipschitz function over $\D$ with constant $L$, that is for all $x_1,x_2\in\D$, we have
\begin{equation}
|f(x_1)-f(x_2)|\leq L\|x_1-x_2\|_2.
\nonumber
\end{equation}
Notice that if the function is not Lipschitz, then there is no hope that we can find the global optimum of $f(\cdot)$ even with infinitely countable evaluations. Thus, the Lipschitz continuity assumption is not a strong assumption. Moreover, functions with larger $L$ are harder to optimize since they change more abruptly over the space.

For any point $x\in\D$, let $r_x = \frac{M-f(x)}{L}$ be the associated radius to the point $x$. By Lipschitz continuity assumption, we know that $x^*\notin \S(x,r_x)$, where, $\S(x,r_x)$ is the set of all points inside the sphere (or circle) with radius $r_x$ centered at $x$ (and single point $x$ if $r_x\leq 0$); otherwise, the Lipschitz assumption is violated. This means if we have a sample at point $x$, then we do not need any more samples inside $\S(x,r_x)$.

The expected value of $r_x$ satisfies $\E[r_x]=\frac{\left|M-\mu_x\right|}{L}$. Since $f(x)$ is a normal random variable $\N(\mu_x,\sigma^2_x)$, using Hoeffding inequality for all $\epsilon>0$, we have
\begin{equation}
\begin{aligned}
\P\left[r_x\,<\,\frac{\left|M-\mu_x\right|}{L}-\epsilon\right]&\leq \exp\left(-\frac{2\epsilon^2L^2}{\sigma_x^2}\right).
\end{aligned}
\nonumber
\end{equation}
Replacing $\epsilon$ with $1.5\frac{\sigma_x}{L}$, the above inequality entails that with high probability (~99\%), $r_x\geq \frac{\left|M-\mu_x\right|-1.5\sigma_x}{L}$. Hence, a ``good" algorithm for exploration should try to find $x$ that maximizes the lower bound on $r_x$. This choice of $x$ will remove a large volume of points from the search space. Note, however, if $x$ is close to the boundaries of $\D$, then it might be the case that most of the volume of the sphere lies outside $\D$. Also, the sphere associated with $x$ might have significant overlap with spheres of other points that are already selected. To fix this issue, we pick the point whose sphere has the largest intersection with unexplored search space in terms of its volume. The pseudo code of this method is described in Algorithm~\ref{alg:explore}, which we refer to as the Next Best exploRative Sample (NBRS) algorithm. NBRS achieves the optimal exploration in the sense that it maximizes the \emph{expected explored} volume.

The value of $\left|M-\mu_x\right|-1.5\sigma_x$ might be negative, especially for large values of $\sigma_x$. This artifact happens at points $x$ that are ``far" from previously observed samples. To prevent/minimize this, we need to make sure that the observed samples affect the mean and variance of all points in the space. For example, if we use the Gaussian kernel $k(x_1,x_2)=\exp(-\frac{1}{\ell_r}\|x_1-x_2\|_2^2)$ for exploration, then we need to choose $\ell_r$ large enough to make sure each observed sample affects all the points in the space, e.g., $\ell_r\geq\sum_{i=1}^d(b_i-a_i)^2$. If we pick small $\ell_r$, then the exploration algorithm starts exploring around the previous samples and extend the explored area gradually to reach to the other side of the search space. This strategy is not optimal if we have limited samples for exploration.

To implement NBRS, we need to maximize the volume
\begin{equation}
g(x) = \textbf{Vol}\left(\D_t\,\cap\,\S\left(x,\frac{\left|M-\mu_{x|\O}\right|-1.5\sigma_{x|\O}}{L}\right)\right)
\nonumber
\end{equation}
where $\D_t$ represents the current unexplored input space.
To evaluate $g(x)$, we take a large number of points $N$ inside the sphere $\S(x,\frac{\left|M-\mu_{x|\O}\right|-1.5\sigma_{x|\O}}{L})$ uniformly at random. Then, for each point, we check if it crosses the borders $[a_i,b_i]$ or falls into the spheres of previously observed samples. If not, we count that point as a newly explored point. Finally, if there are $n$ newly explored points, then we set $g(x)\approx\frac{n}{N}\left(\frac{\left|M-\mu_{x|\O}\right|-1.5\sigma_{x|\O}}{L}\right)^{\!\!d}$.

To optimize $g(x)$, one can use deterministic and derivative free optimizers like DIRECT \cite{Jones93}. The problem is that DIRECT only optimizes Lipschitz continuous functions; however, $g(x)$ is not necessarily Lipschitz continuous. In our implementation, we take a large number of points inside $\D_t$ and evaluate $g(\cdot)$ at those points and pick the maximum. This method might be slower than DIRECT, but avoids inaccurate results of DIRECT especially when $\D_t$ describes a small region.

\subsection{Exploitation}

In the exploitation phase of the algorithm, we would like to use the information gained in the exploration phase to find the optimal point of $f(\cdot)$. Suppose we have explored the search space with $t$ samples and we want to find $x^*\in\D_t$. In order to exploit, we would like to find points $x$ whose sphere is small. The reason is that if $r_x=\frac{M-f(x)}{L}\leq\gamma$ is small enough, then by \emph{local} strong convexity of $f(\cdot)$ around $x^*$, for some constant $\kappa$ we have
\begin{equation}
\frac{\kappa}{2}\|x-x^*\|_2^2\leq M-f(x)\leq L\gamma.
\nonumber
\end{equation}
Following the argument in Section 3.1, we estimate $r_x$ by its mean $\E[r_x]=\frac{|M-\mu_x|}{L}$. By Hoeffding inequality, for all $\epsilon>0$, we have
\begin{equation}
\begin{aligned}
\P\left[r_x\,>\,\frac{\left|M-\mu_x\right|}{L}+\epsilon\right]&\leq \exp\left(-\frac{2\epsilon^2L^2}{\sigma_x^2}\right).
\end{aligned}
\nonumber
\end{equation}
Similarly, replacing $\epsilon$ with $1.5 \frac{\sigma_x}{L}$, the above inequality entails that with high probability (~99\%), $r_x\leq \frac{\left|M-\mu_x\right|+1.5\sigma_x}{L}$. Hence, a ``good" algorithm for exploitation should try to find the point $x$ that minimizes the upper bound on $r_x$. This choice of $x$ introduces the expected closest point to $x^*$. We present the pseudo code of this method in Algorithm~\ref{alg:exploit}.

\begin{algorithm}[t]
\caption{Next Best exploItive Sample (NBIS)}
\label{alg:exploit}
\begin{algorithmic}
\STATE {\bf Input}: Maximum $M$, Lipschitz Constant $L$ and Set of observed samples $\{(x_1,f(x_1)),\ldots,(x_q,f(x_q))\}$
\STATE {\bf Output}: Next best exploitive sample $x$
\STATE
\STATE $\displaystyle\D_q = \D - \bigcup_{i=1}^q\S(x_i,r_{x_i})$
\STATE
\STATE $\displaystyle x\longleftarrow\argmin_{x\in\D_q}\; \textbf{Vol}\left(\S\left(x,\frac{\left|M-\mu_{x|\O}\right|+1.5\sigma_{x|\O}}{L}\right)\right)$
\STATE
\STATE
\end{algorithmic}
\end{algorithm}

The optimization in Algorithm~\ref{alg:exploit} is nothing but minimizing
\begin{equation}
h(x) = \frac{\left|M-\mu_{x|\O}\right|+1.5\sigma_{x|\O}}{L}.
\nonumber
\end{equation}
To optimize $h(x)$, again we take a large number of points in $\D_q$ (the current unexplored space) uniformly at random and evaluate $h(\cdot)$ on those and pick the minimum.

\subsection{Exploration-Exploitation Trade-off}

The main algorithm consists of an initial exploration phase followed by exploitation. Notice that we are using GP as an estimate of the unknown function and our method, like EI, highly relies on the quality of this estimation. On a high level, if the function is very complex, i.e., has large Lipschitz constant $L$, then we need more exploration to fit better with GP. Small values of $L$ correspond to flatter functions that are easier to optimize. Thus, in general, we expect the number of exploration steps to scale up with $L$. As a rule of thumb, functions we normally deal with satisfy $2<L<20$, for which we spend 20\% of our budget in exploration and the rest in exploitation.

We use different kernel widths for the exploration and exploitation phases. In the case of exploration for complex functions, if we have enough budget (and hence, enough explorative samples), the kernel width can be set to a small value to fit a better local GP model. However, if we do not have enough budget, we need to take the kernel width to be large. In the case of exploitation, we pick the kernel width under which EI achieves its best performance.

Note that the choice of $M$ and $L$ plays a crucial role in this algorithm. If we pick $L$ larger than the true Lipschitz function, then the radius of our spheres shrink and hence we might need more budget to achieve a certain performance. Choosing $L$ smaller than the true Lipschitz is dangerous since it makes the spheres large and increases the chance of including the optimal point in a sphere and hence removing it. Thus, it is better to choose $L$ slightly larger than our estimate of the true Lipschitz to be on the safe side.

The method is less sensitive to the choice of $M$, since the derivative of the radius with respect to $M$ is proportional to $\frac{1}{L}$. Thus, as long as we do not over estimate $M$ significantly, the $\frac{1}{L}$ factor prevents the spheres to become very large (and include/remove the optimal point). Small values of $M$, make the spheres smaller and hence, if we underestimate $M$, we would need more budget to achieve certain performance. However, if $M$ is significantly (proportional to $L$) smaller than the true maximum of the function, then the algorithm will look for the point that achieves $M$ and hence will perform poorly.

\section{Experimental Results} \label{sec:exper}
\label{sec:results}
\begin{figure}[t]
\label{fig:contour}
\begin{center}
\begin{tabular}{cc}
\includegraphics[width=1.44in,height=1.2in]{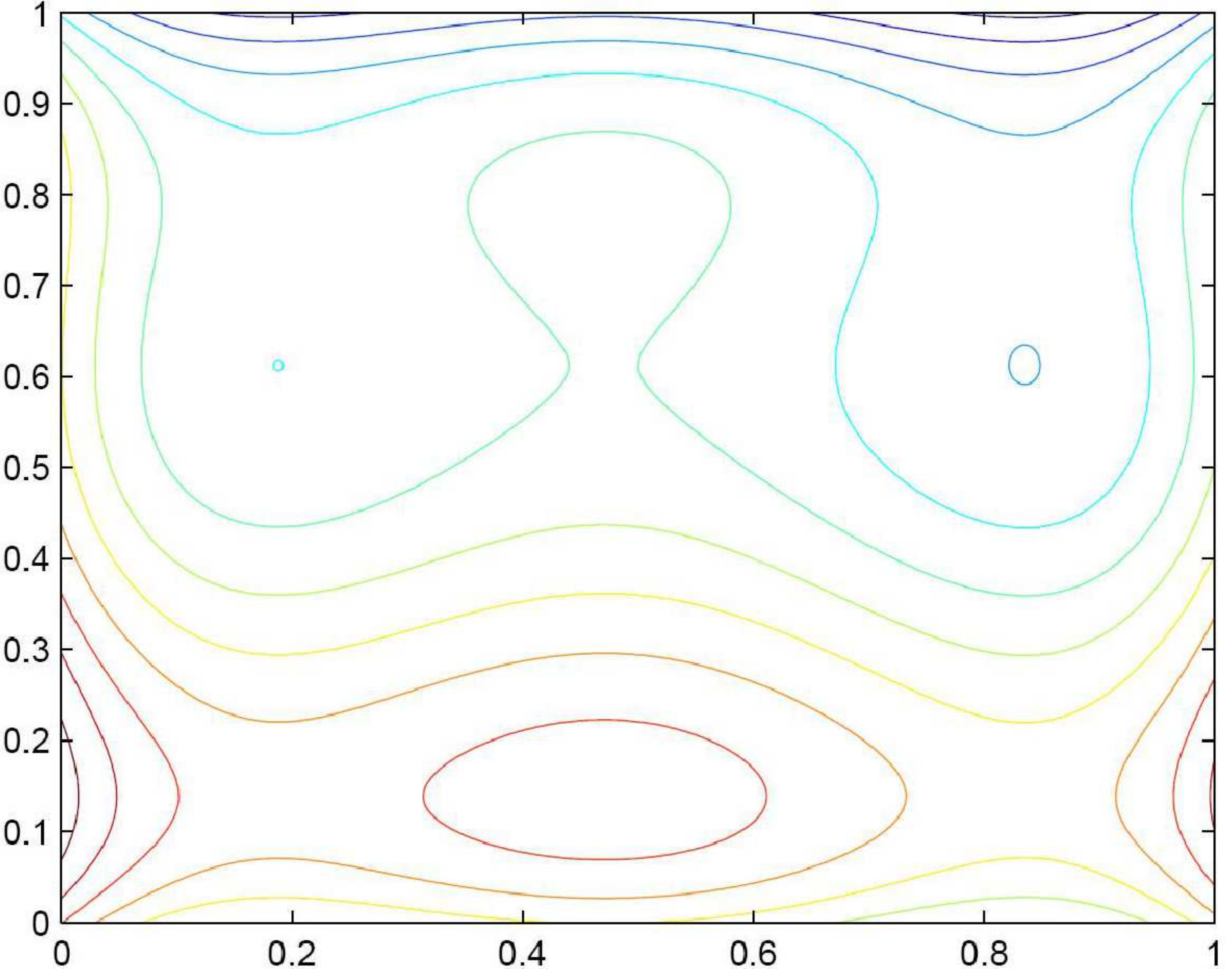}&
\includegraphics[width=1.44in,height=1.2in]{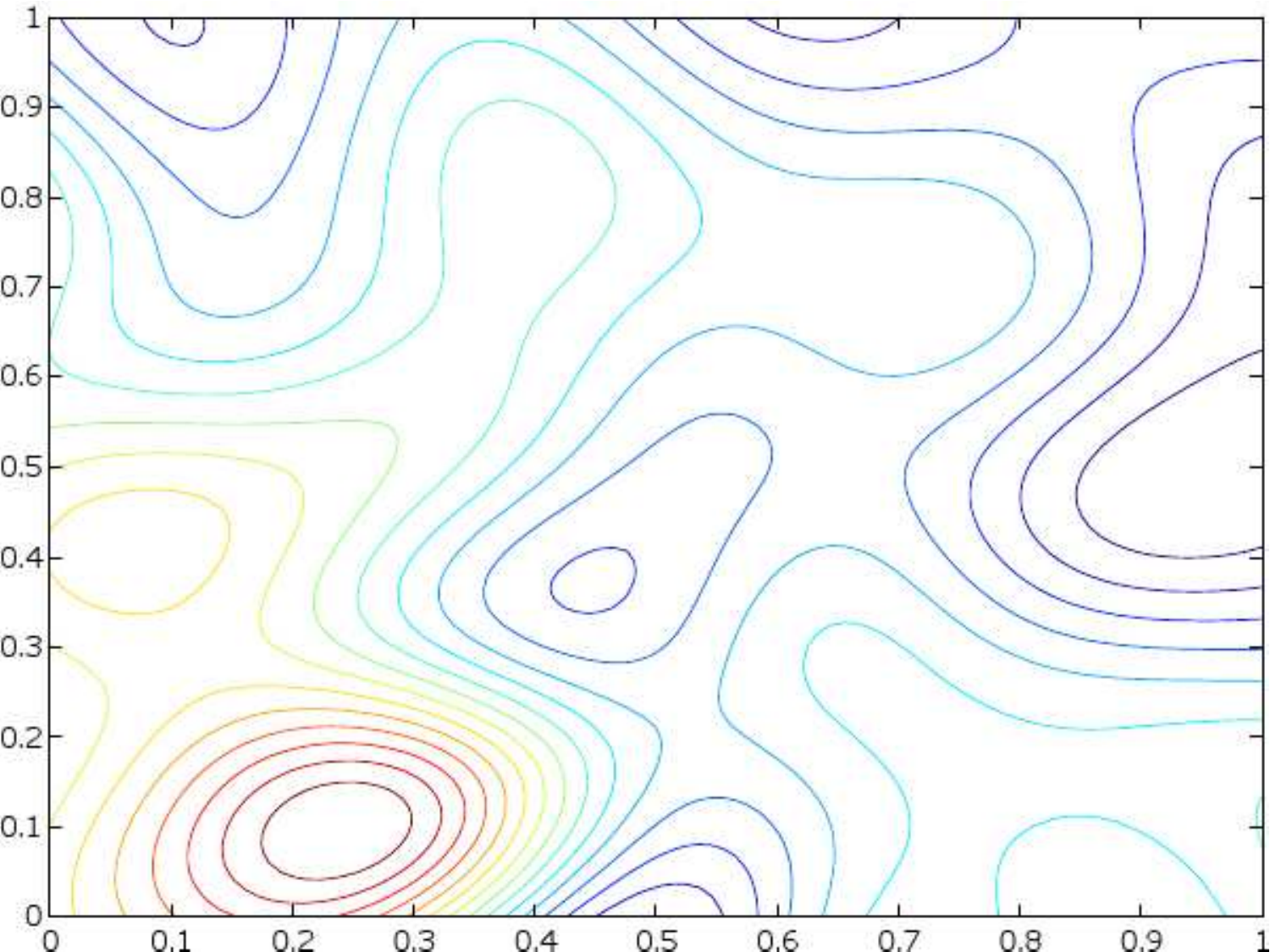}\\
Fuel Cell & Hydrogen\\
\includegraphics[width=1.44in,height=1.2in]{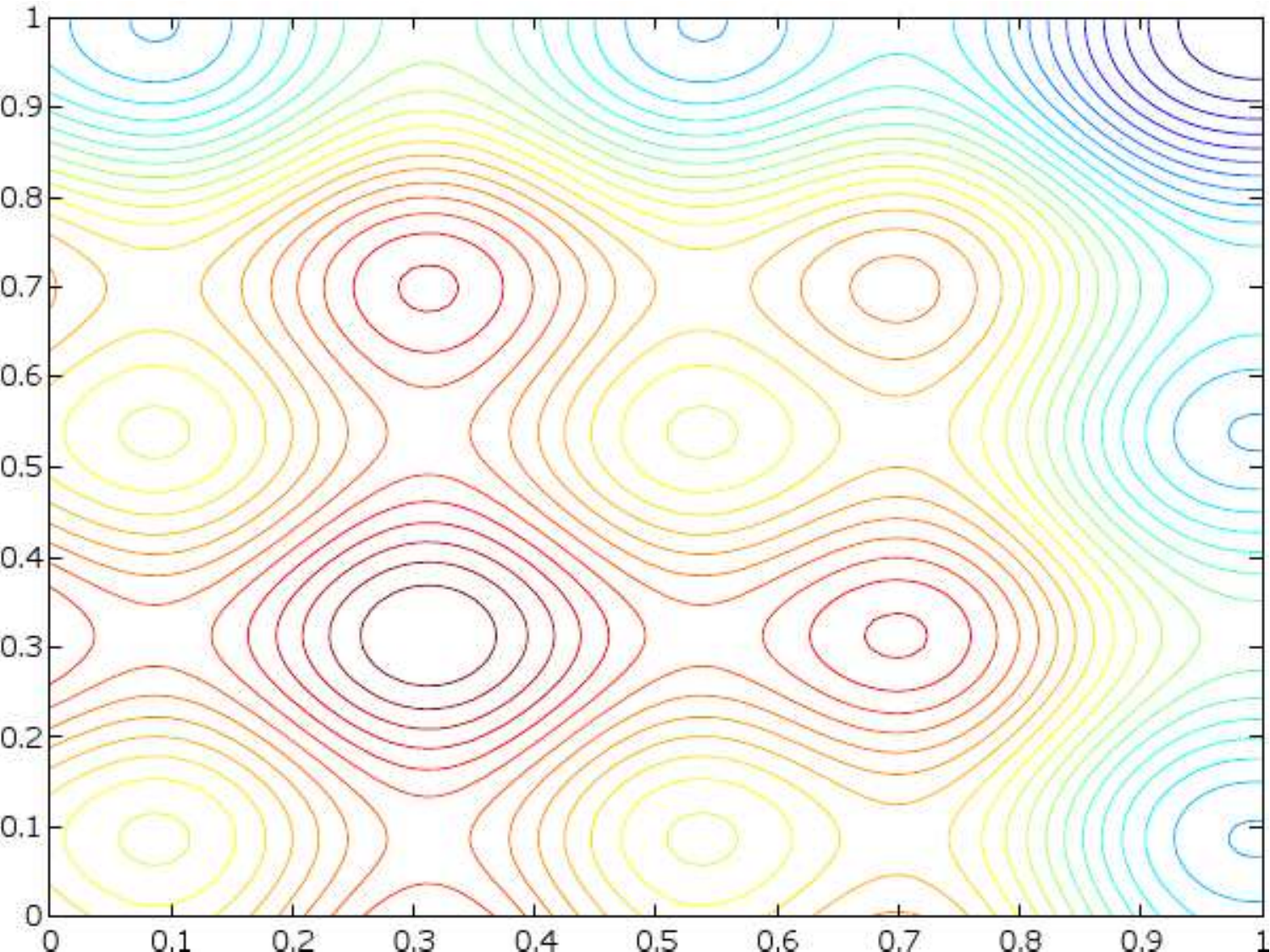}&
\includegraphics[width=1.44in,height=1.2in]{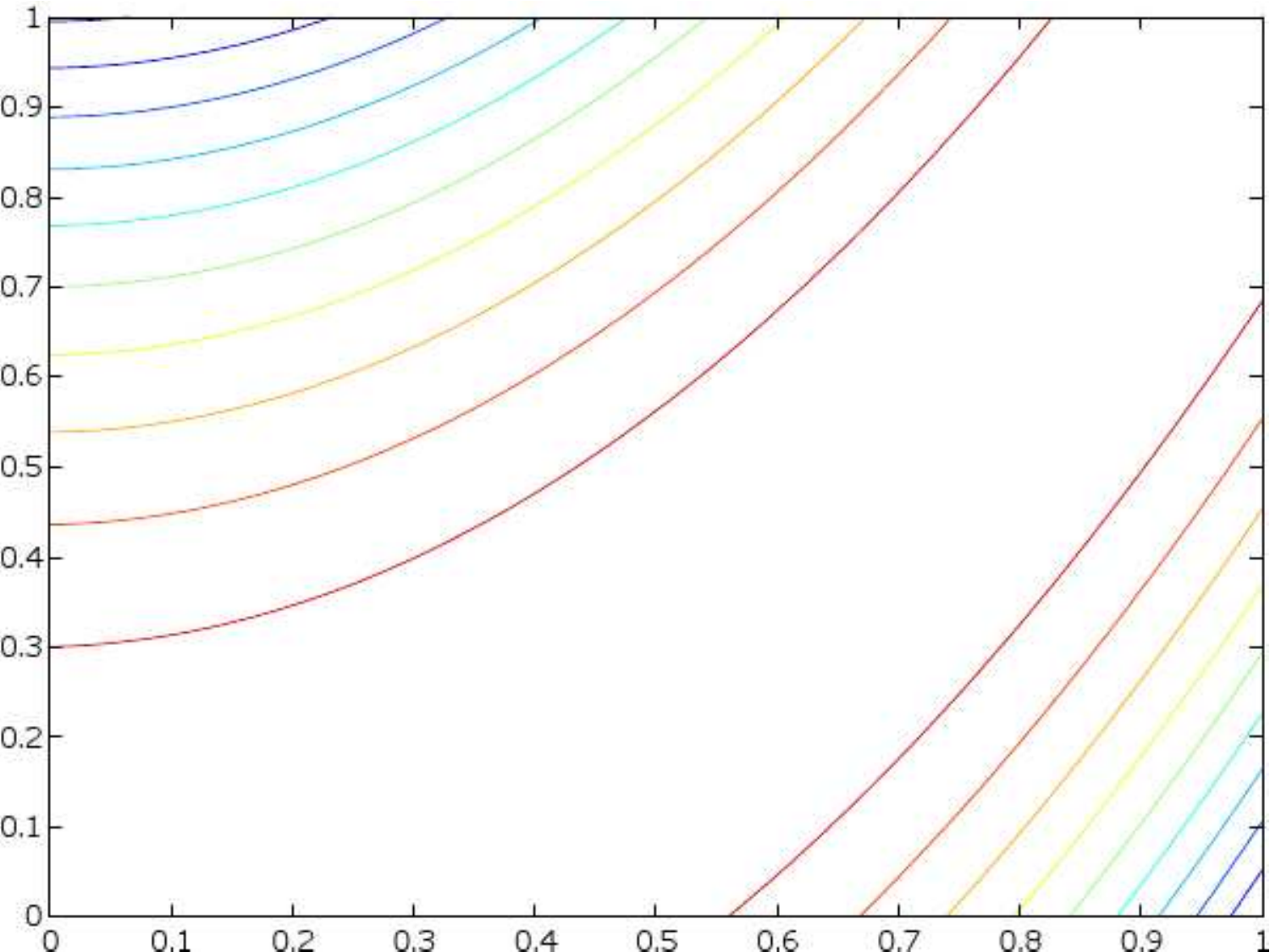}\\
Cosines& Rosenbrock\\
\end{tabular}
\end{center}
\caption{The contour plots for the four $2-$dimension proposed benchmarks.}
\end{figure}

\begin{table*}[t]
\begin{center}
\makebox[\textwidth][c]{
\footnotesize{
\begin{tabular}{|l|c||l|c|}
\hline
\multirow{2}{*}{Cosines(2)} & \footnotesize $1\!-(u^2\!+v^2\!-0.3\cos(3\pi u)\!-0.3\cos(3\pi v))$ & \multirow{2}{*}{Rosenbrock(2)} & \multirow{2}{*}{\footnotesize$10\!-\!100(y\!-x^2)^2\!\!-\!(1\!-x)^2$}\\
&\footnotesize $u=1.6x-0.5, v=1.6y-0.5$ & & \\\hline
\multirow{2}{*}{Hartman(3,6)} & \footnotesize$\sum_{i=1}^4 \Omega_i \exp\left(-\!\sum_{j=1}^d A_{ij}(x_j-P_{ij})^2\right)$ & \multirow{2}{*}{Michalewicz(5)}& \multirow{2}{*}{\footnotesize$-\sum_{i=1}^{5}\sin(x_i)\sin\left(\frac{i\, x_i^2}{\pi}\right)^{\!\!20}$}\\
 & \footnotesize $\Omega_{1\times 4}, \; A_{4 \times d},\; P_{4\times d}$ are constants &  & \\\hline
Shekel(4)& \multicolumn{3}{c|}{ \footnotesize$\sum_{i=1}^{10}\frac{1}{\omega_i+\Sigma_{j=1}^4(x_j-B_{ji})^2}$ \qquad $\omega_{1\times 10}, \; B_{4a \times 10}$ are constants}\\ \hline
\end{tabular}}}
\end{center}
\caption{Benchmark Functions}\label{tab:functions}
\end{table*}

In this section, we compare our algorithm with EI under different scenarios for different functions. We consider  six well-known synthetic benchmark functions:
\begin{itemize}
\item [(1,2)] Cosines \cite{Anderson00} and Rosenbrock \cite{Brunato06} over $[0,1]^2$
\item [(3,4)] Hartman(3,6) \cite{GOP78} over $[0,1]^{3,6}$
\item [(5)] Shekel \cite{GOP78} over $[3,6]^4$
\item [(6)] Michalewicz \cite{Michal94} over $[0,\pi]^5$
\end{itemize}
The mathematical expression of these functions are shown in Table 1. Moreover, we use two benchmarks derived from real-world applications:
\begin{itemize}
\item [(1)] Hydrogen \cite{hydrogen} over $[0,1]^2$
\item [(2)] Fuel Cell \cite{azimi10AAAI} over $[0,1]^2$
\end{itemize}
The contour plots of these two benchmarks along with the Cosines and Rosenbrock benchmarks are shown in Fig~\ref{fig:contour}. The Fuel Cell benchmark is based on optimizing electricity output of microbial fuel cell by modifying some nano structure properties of the anodes. In particular, the inputs that we try to adjust are the average area and average circularity of the nano tube and the output that we try to maximize is the power output of the fuel cell. We fit a regression model on a set of observed samples to simulate the underlying function $f(\cdot)$ for evaluation. The Hydrogen benchmark is based on maximizing the Hydrogen production of a particular bacteria by varying the PH and Nitrogen levels of its growth medium. A GP is fitted to a set of observed samples to simulate the underlying function $f(\cdot)$. We consider a Lipschitz constant $L\approx 3$ for all of the benchmarks, except for Cosines and Michalewicz with $L\approx 6$ and Rosenbrock with $L\approx 45$. For the sake of comparison, we consider the normalized versions of all these functions and hence $M=1$ in all cases. As mentioned previously, we spend 20\% of the budget on exploration and 80\% on exploitation.

\begin{table*}[t]
\label{table:result}
\caption{Comparison of the best results of EI, NBRS+EI and NBRS+NBIS. This result shows that our algorithm outperforms the other two counterparts significantly in most cases both in terms of the mean and variance of the performance.}
\begin{center}
\begin{tabular}{|l|  c  c  c  c |}
\hline
 &  $\quad\qquad$EI$\quad\qquad$ &  $\quad\qquad$EI$_M$ $\quad\qquad$ & $\quad${NBRS+EI}$\quad$ &$\quad${NBRS+NBIS}$\quad$    \\ \hline
{Cosines}$\qquad$ &    $.0736\pm .016$  & $.2938 \pm .020$  & $.1057\pm .029  $ &  $\mathbf{.0270\pm .009}$\\ \hline
{Fuel Cell} &  $.1366\pm .006$  &  $.2232\pm.007 $& $.1357\pm .004 $ &   $\mathbf{.0965\pm .004}$  \\ \hline
{Hydrogen} & $.0902\pm .004$  & $.1689\pm .012 $& $.1149\pm  .004$   & $\mathbf{.0475\pm .006}$ \\ \hline
{Rosen}&   $.0134\pm .001 $  & $.0153 \pm .003 $  & $.0163 \pm .001$  &  $\mathbf{.0034 \pm .000}$ \\  \hline
{Hart(3)}&     $.0618\pm .006$ & $.0837 \pm .001$ &  $.0450\pm  .003$   &  $\mathbf{.0384\pm  .003}$ \\ \hline
{Shekel} &     $.3102 \pm .017$  & $.4104 \pm .021 $& $\mathbf{.3011\pm .018}$   &  $.3240\pm .030$\\ \hline
{Michal}& $.5173\pm .010$  & $.5210 \pm .008$&   $.5011\pm .010$  &  $\mathbf{.4554\pm .019}$\\ \hline
 {Hart(6)}&     $.1212\pm .002$   & $.2207\pm .006$&  $.1235\pm .002$  &  $\mathbf{.1020\pm .003}$ \\  \hline
\end{tabular}
\end{center}

\end{table*}

\subsection{Comparison to EI}
In the first set of experiments, we would like to compare our algorithm with the best possible performance of EI. For each benchmark, we search over different values of the kernel width and find the one that optimizes EI's performance. Fig.~\ref{fig:1} is plotted using these optimal kernel widths and shows that the best performance of EI happens when we take only one random sample from a given budget. This performance is then used as the baseline for comparison in Table 2. In addition to EI, we introduced a new version of EI, called EI$_M$. Instead of taking the expectation of improvement $I$ from $0$ to infinity, (equation \ref{eq:ei}), we calculate the expectation of improvement from $0$ to $M-y_{max}$ assuming $M$ is given. This simple change decreases the level of exploration of EI and changes its behavior to be more exploitative than explorative.  Using GP as our posterior model, the following lemma represents the EI$_M$. The proof is in supplementary document. 

\begin{lemma}
Let $u_1=\frac{y_{max}-\mu_x}{\sigma_x}$ and $
u_2=\frac{M-\mu_x}{\sigma_x}$, then 

\begin{equation}
\begin{aligned}
EI_M(x) &= \mathbb{E}\left[(f(x)-y_{\max})\,\mathbb{I}_{\{0\leq f(x)-y_{\max} \leq M-y_{max}\}}\right]\\
&=\sigma(x)\big(-u_1\Phi(u_2)+u_1\Phi(u_1)+\phi(u_1)\big).
\end{aligned}
\end{equation} 
\end{lemma}

In light of the results of Fig.~\ref{fig:1}, we are also interested in whether our exploration algorithm can be used to improve the performance of EI. To this end, we replace the proposed exploitation algorithm with EI to examine if our exploration strategy helps EI. We refer to this setting as NBRS+EI.

Table 2 summarizes the mean and variance of the performance, measured as the ``Regret"$=M-\max f(x_{\O})$, for different benchmarks estimated over $1000$ random runs. Interestingly, EI can consistently outperform the EI$_M$ in all benchmarks. This shows that decreasing the exploration rate of EI could degrade the performance. 

It is easy to see that in all benchmarks, our algorithm (NBRS+NBIS) outperforms EI consistently except for the Shekel benchmark where EI and NBRS+EI have slightly better performances. We suspect this is due to the fact that we have not optimized our kernel widths, where as the EI kernel width is optimized.

We also note that NBRS+EI does not lead to any consistent improvement over EI. This is possibly due to the fact that EI does not take advantage of the reduced search space produced by NBRS during selection. 


\begin{figure*}[t]
\begin{center}
\makebox[\textwidth][c]{
\addtolength{\tabcolsep}{-0.7em}
\begin{tabular}{c c c c}
\includegraphics[width=1.7in, height=1.45in]{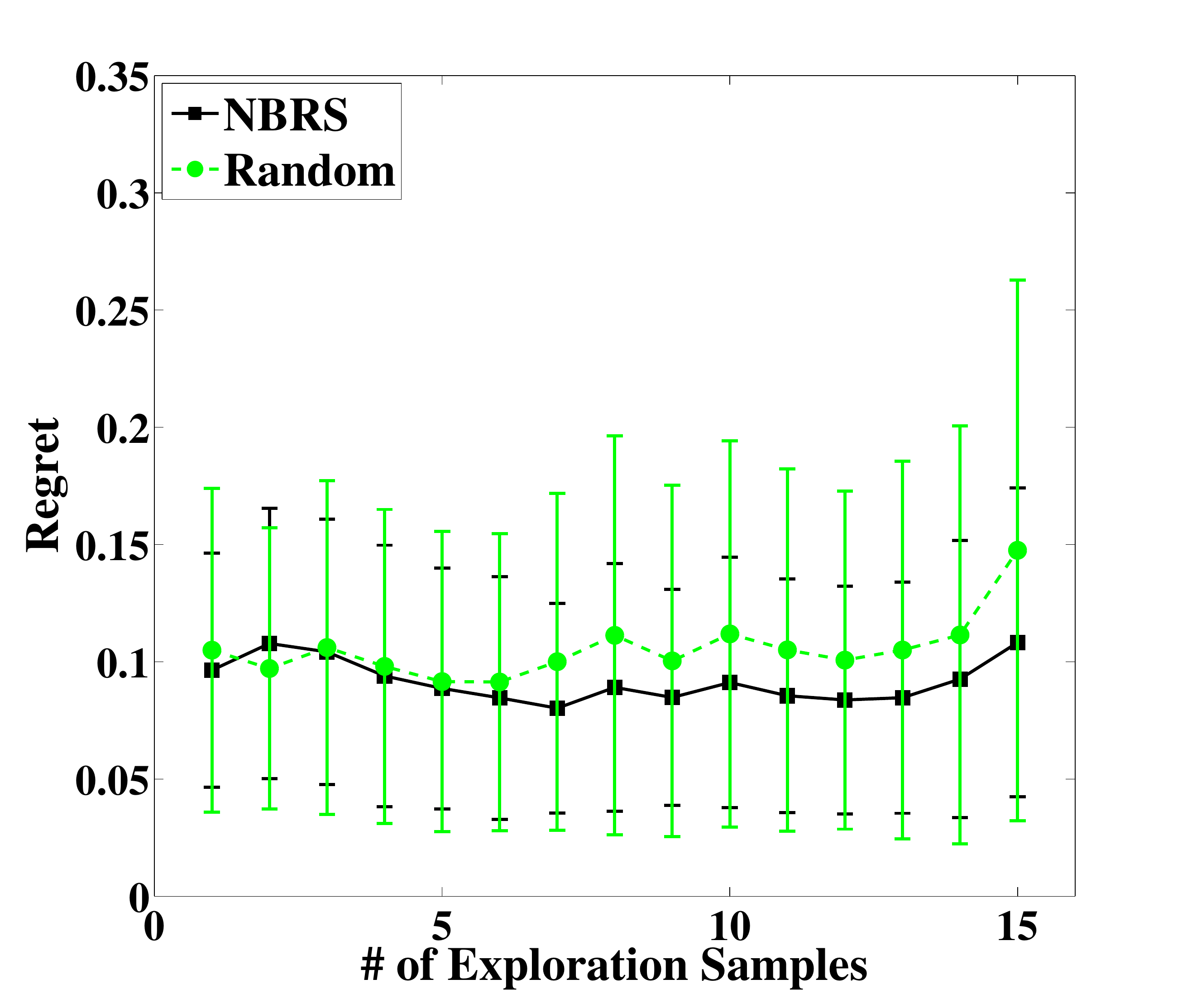}&
\includegraphics[width=1.7in, height=1.45in]{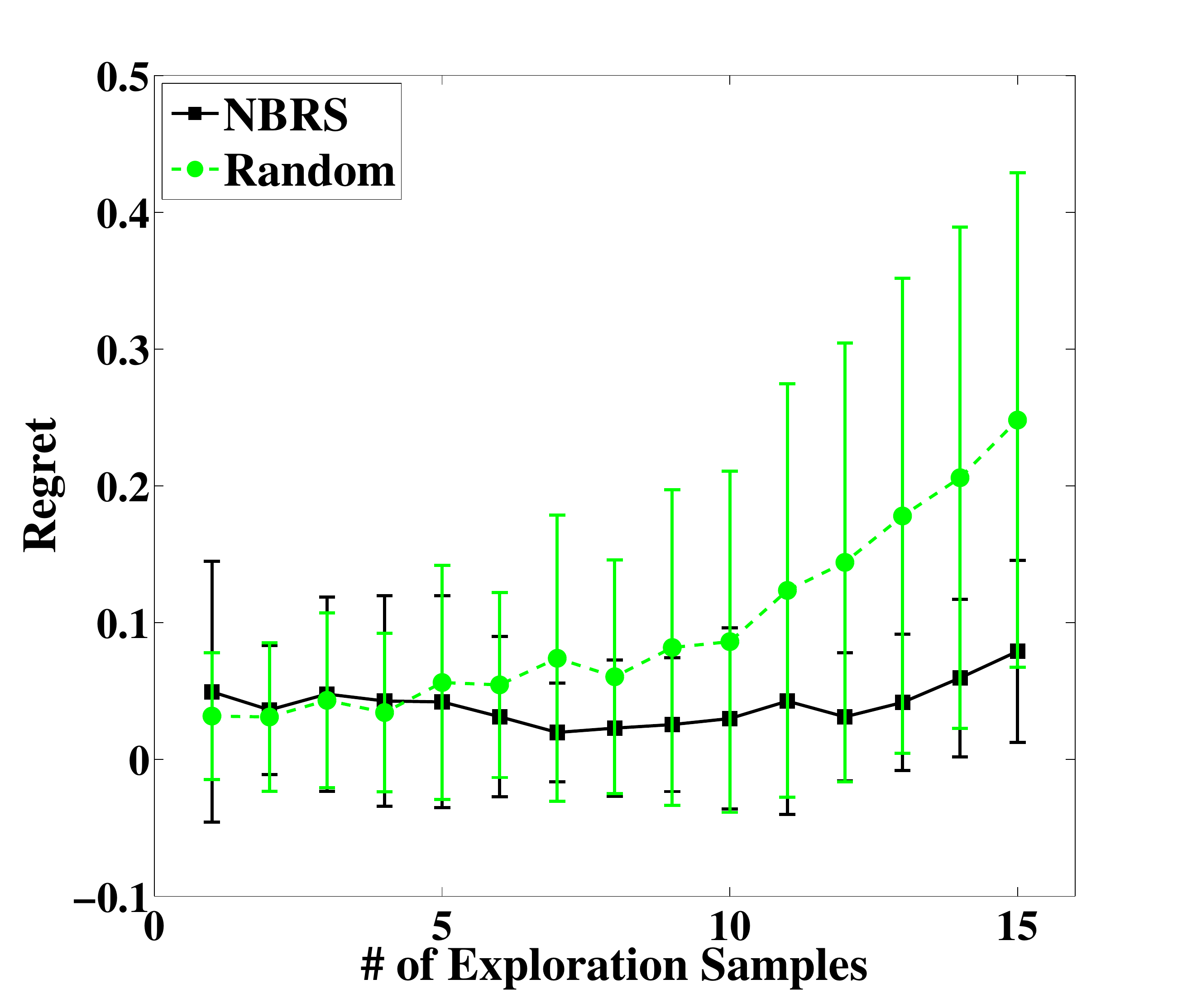}&
\includegraphics[width=1.7in, height=1.45in]{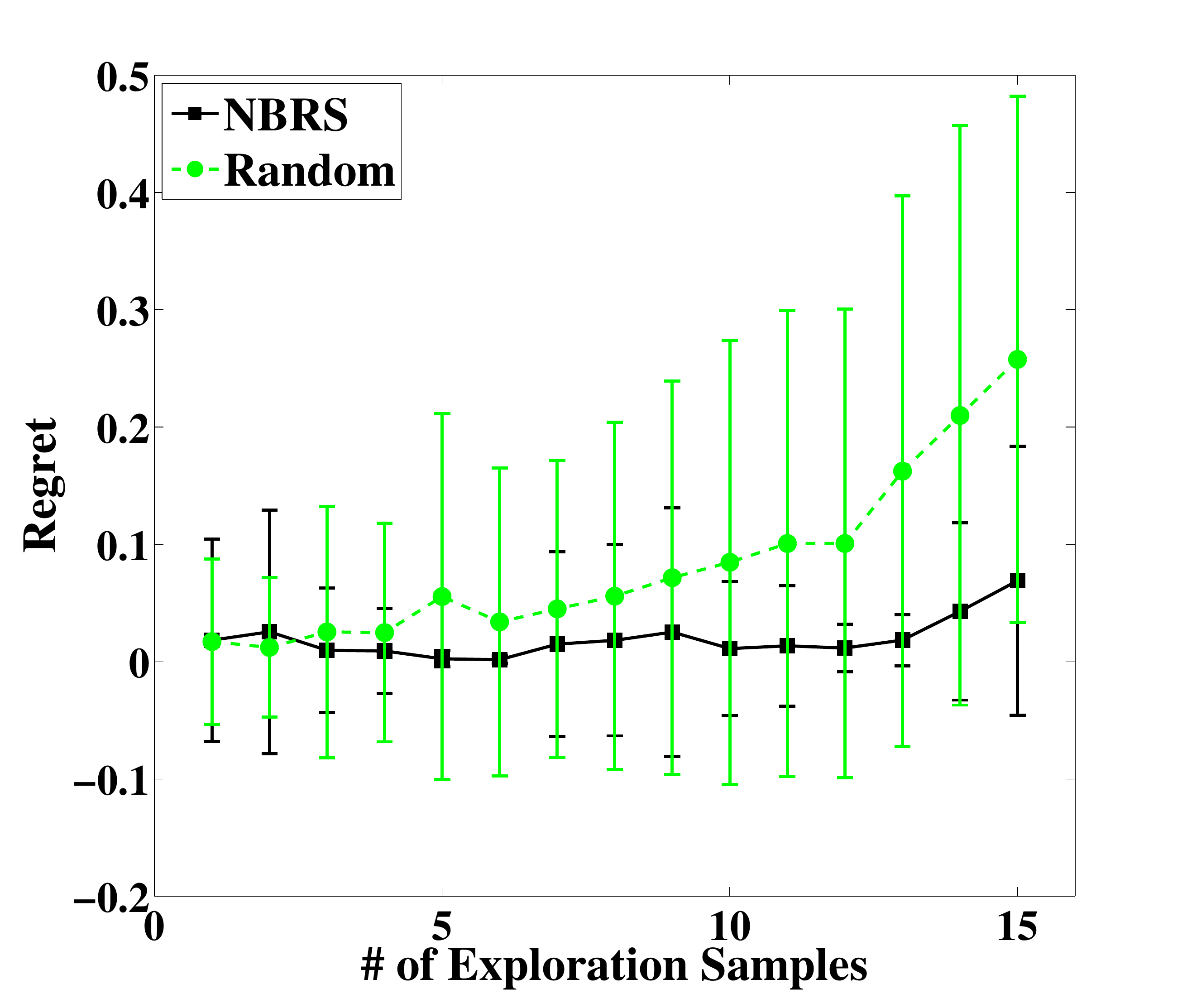}&
\includegraphics[width=1.7in, height=1.45in]{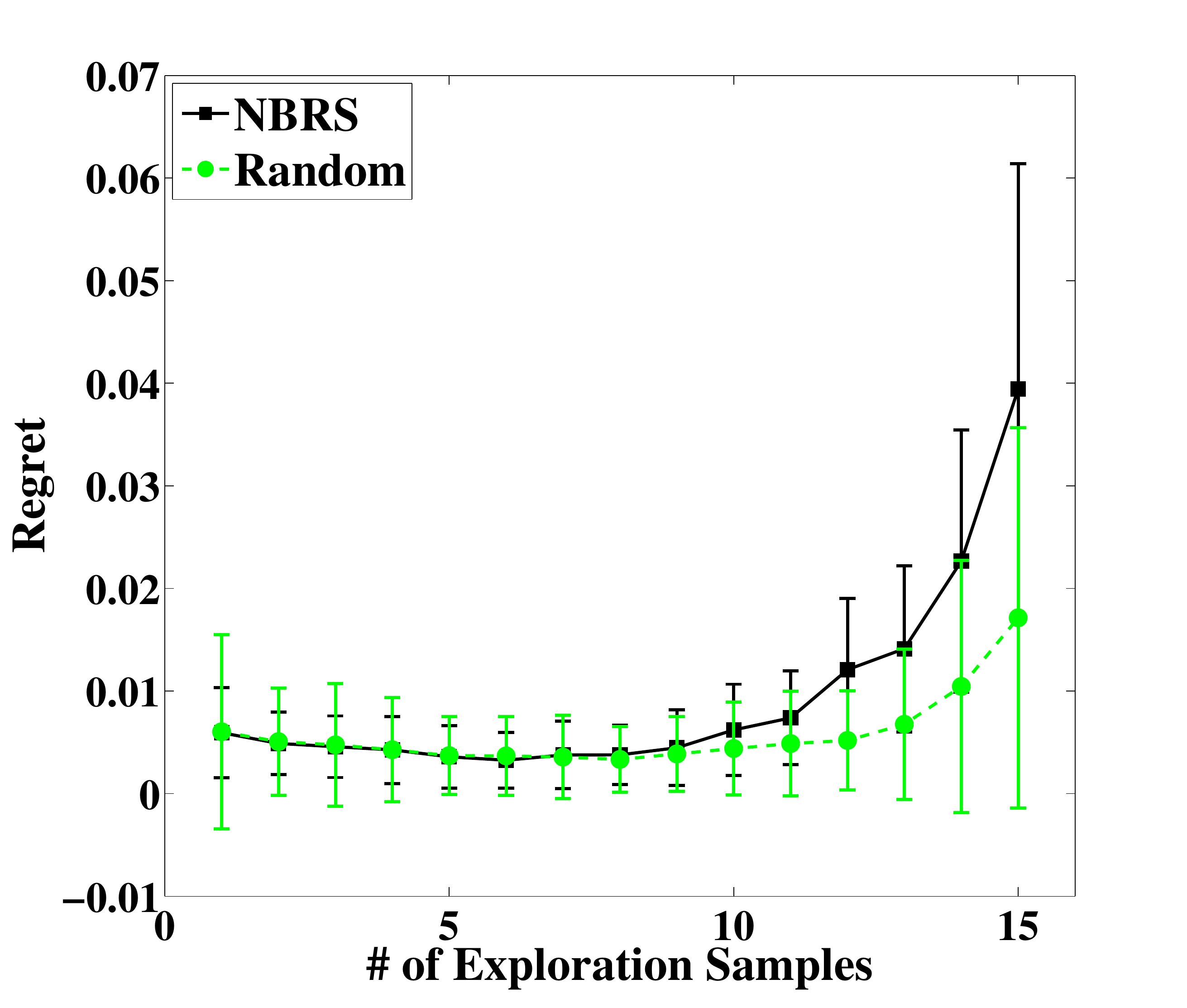}\\
Fuel Cell & Hydrogen & Cosines & Rosenbrock\\
\includegraphics[width=1.7in, height=1.45in]{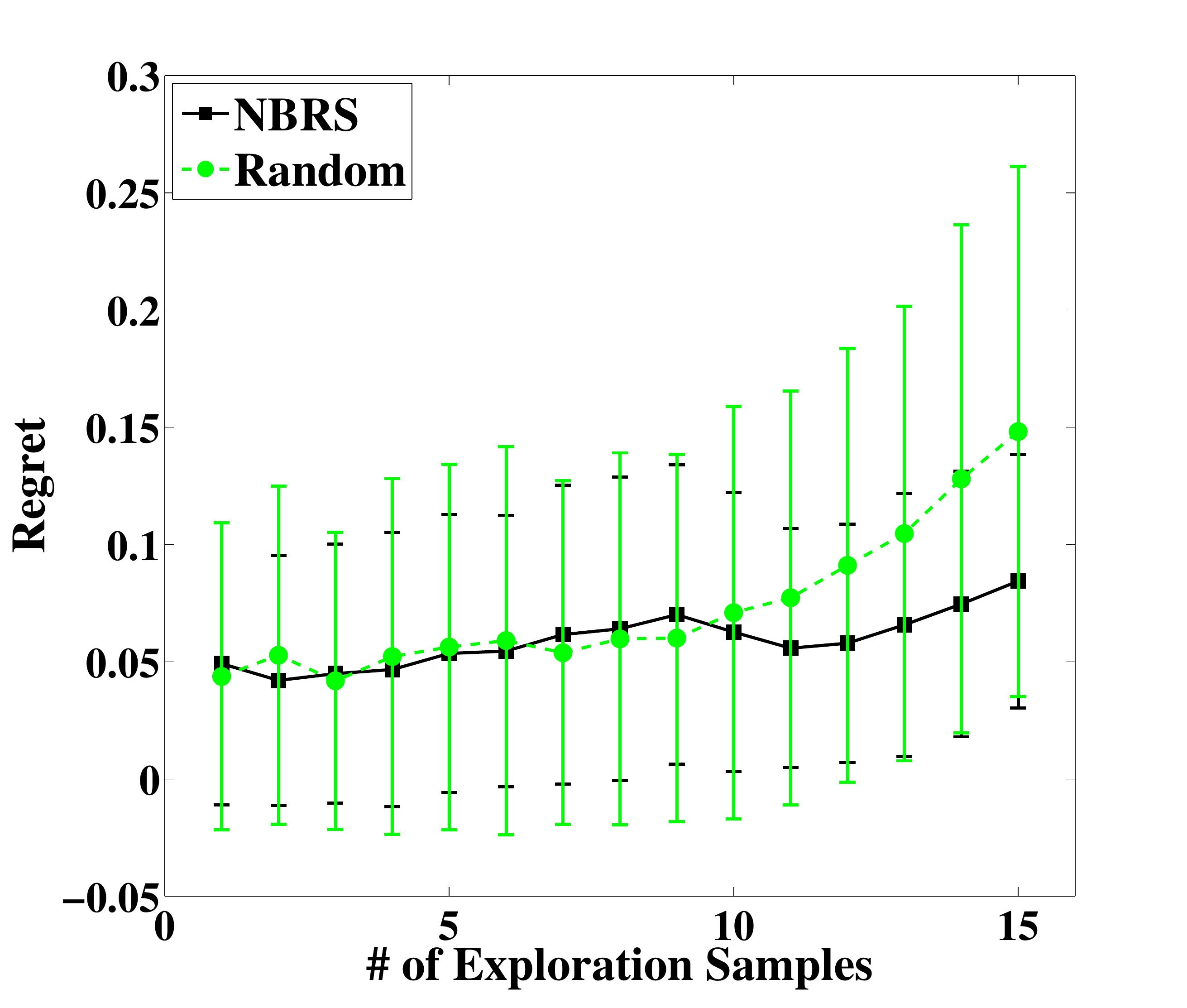}&
\includegraphics[width=1.7in, height=1.45in]{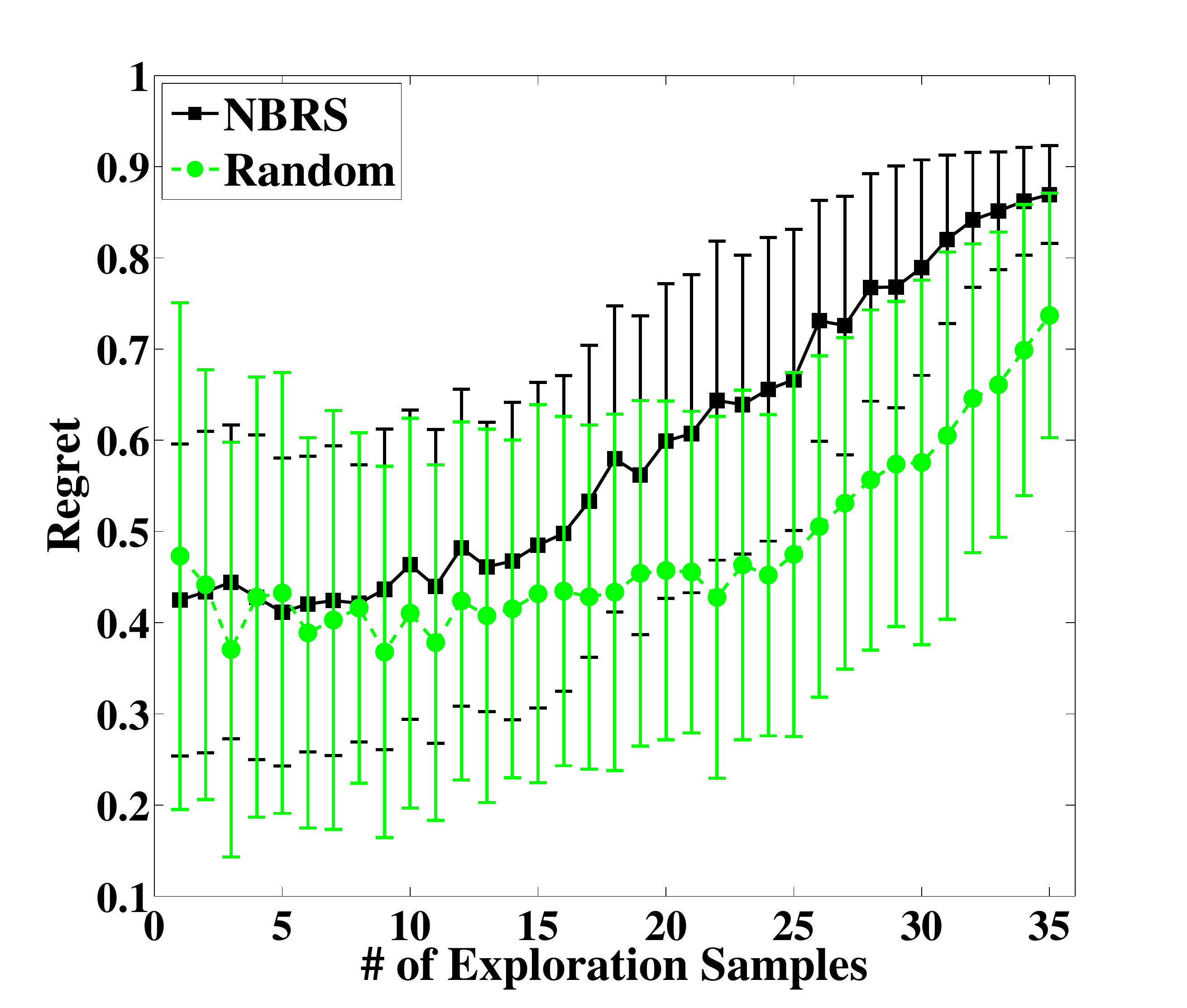}&
\includegraphics[width=1.7in, height=1.45in]{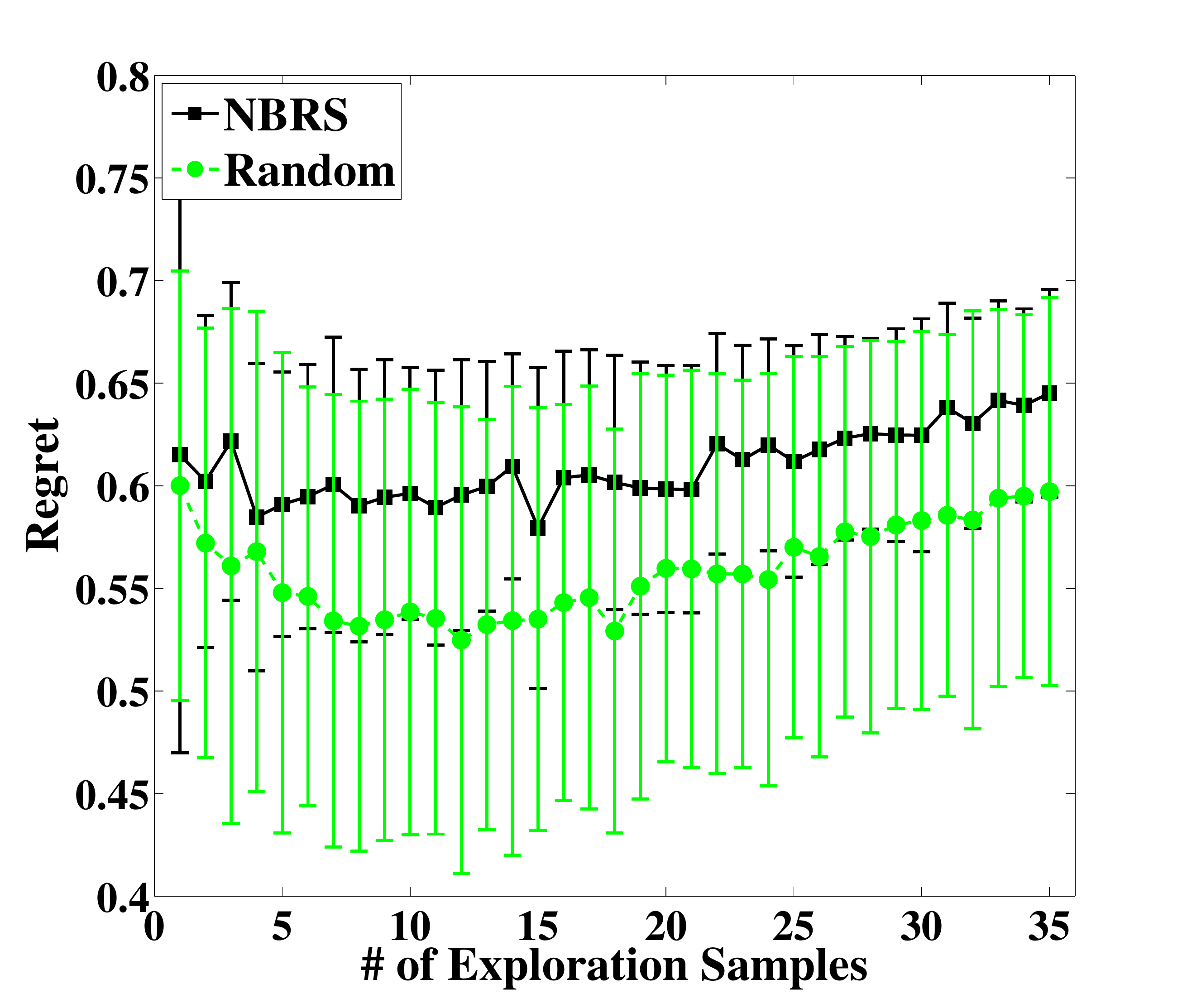}&
\includegraphics[width=1.7in, height=1.45in]{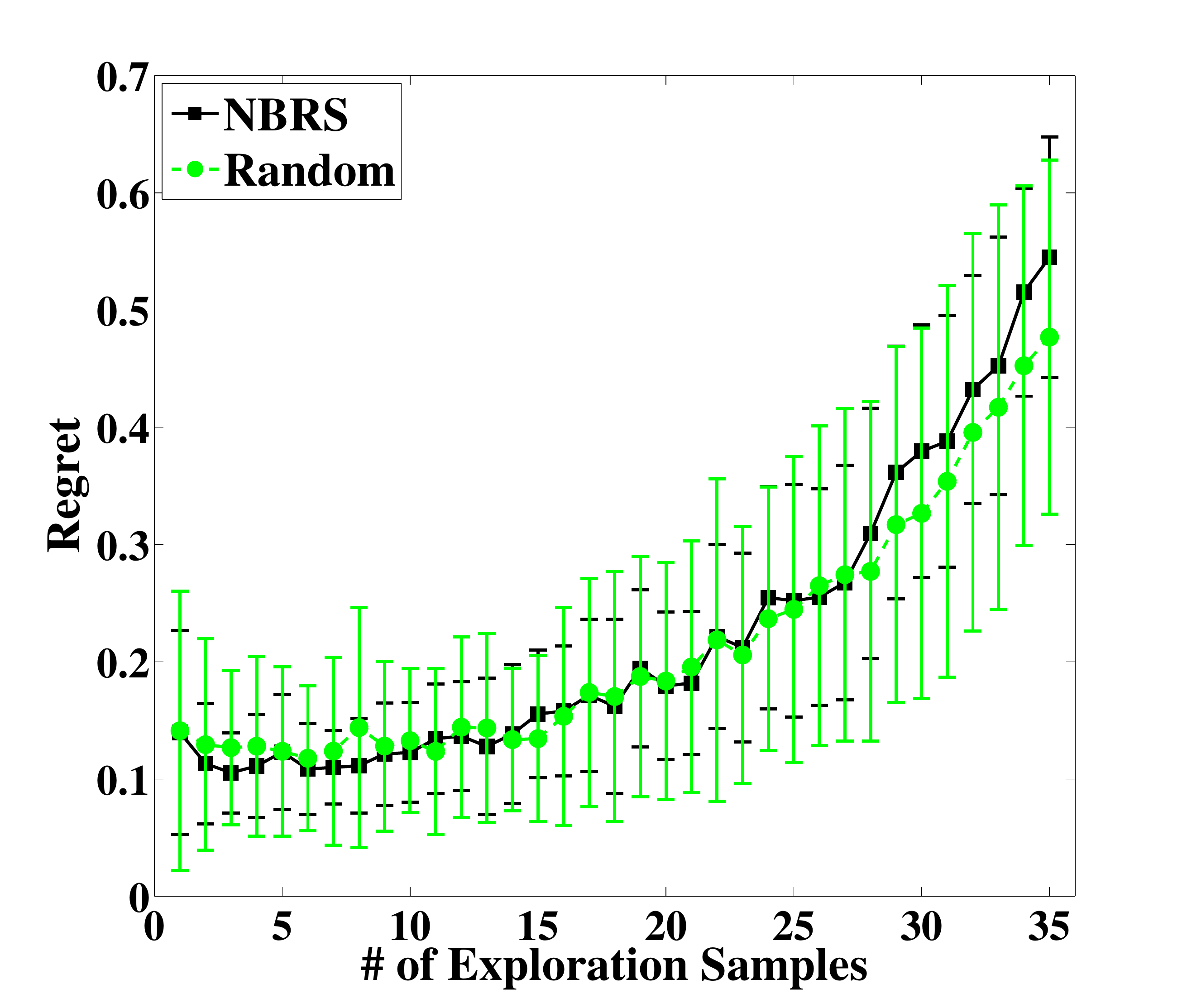}\\
Hartman(3) & Shekel & Michalewicz & Hartman(6)\\
\end{tabular}}
\end{center}
\caption{Plot of regret versus the number of explorations for NBIS algorithm. For a fixed budget $n_b$, we run a number of experiments as follows: first we consider the case where there are $1$ explorative sample (either random or NBRS) followed by $n_b-1$ EI samples, next we consider the case where where there are $2$ explorative samples followed by $n_b-2$ EI samples and so on. For 2D and 3D functions, we let $n_b=15$ and for high-dimensional functions, we let $n_b=35$. This result shows that in most cases, our exploration is a) better than random, and b) necessary, since the regret achieves its minimum somewhere apart from zero. On average, we need to explore 20\% of our budget, however, this portion can be optimized if we consider any specific function. The error bar here is the variance of the regret over different runs. This shows that our regret variance is smaller.}
\label{fig:exp-result}
\end{figure*}

\subsection{Exploration Analysis}
In the second set of experiments, we would like to compare our exploration algorithm NBRS with random exploration when using NBIS for exploitation. As discussed previously, both random exploration and NBRS fail to produce better performance when used with EI. Thus, it is interesting to see whether they can help NBIS in terms of the overall regret, and if so which one is more effective. Figure~\ref{fig:exp-result} summarizes this result for all benchmarks. For a fixed budget $n_b$, we start with $1$ explorative sample (either using NBRS or random) followed by $n_b-1$ NBIS samples; next, we start with $2$ explorative samples followed by $n_b-2$ NBIS samples and so on. In each case, we average the regret over $1000$ runs. The black line corresponds to the NBRS exploration and the green line corresponds to the random exploration. We will discuss each function in more details later, but in general, this result shows that our exploration algorithm is a) better than random exploration and b) necessary. To see why it is necessary, notice that the minimum regret on all curves is achieved for a non-zero number of NBRS samples. This means unlike EI, our exploitation algorithm benefits from NBRS.

Looking closer into the results, we see that NBRS always lead to a smaller regret comparing to the random exploration. On the Shekel benchmark, we see that random exploration has better performance if we spend majority of the budget to \emph{explore}. However, for a \emph{reasonable} amount of exploration that leads to the minimum regret (5 to 10 experiments), random exploration and NBRS achieve similar performance. 

On our $6$-dimensional benchmark Hartman(6), we notice that random exploration and NBRS behave very similarly. This shows that the input space is so large that no matter how clever you explore, you will not likely to improve the performance for the limited budget of $35$.

NBRS starts from an initial point and explores the input space step by step. Imagine you are in a dark room with a torch in your hand and you want to explore the room. You start from an initial point and little by little walk through the space until you explore the whole space. This is exactly how NBRS does the exploration. Roughly speaking, NBRS minimizes $\mu_{x|\O}+1.5\sigma_{x|\O}$ and hence, if a point is far from previous observations, i.e., $\sigma_{x|\O}$ is large, it is unlikely to be chosen. We see this effect in all functions, but most clearly in the Michalewicz benchmark. When the number of explorative samples is smaller than $10$, the step-by-step explore procedure cannot explore the whole space and the exploitation can be trapped in local minima. For $10-15$ explorative samples, NBRS can walk through the entire space fairly well and hence we get a minimum regret. For more than $15$ explorative samples, since the space is well explored, we are wasting the samples that could be potentially used to improve our exploitation and hence, the performance becomes worse.

Finally, this investigation suggests that the result in Table 2 can be further improved by taking different number of explorative samples for different functions. To minimize parameter tuning, we chose to explore 20\% of our budget. In general, this ratio can be adjusted according to the property of the function (e.g., the Lipschitz constant).

\section{Conclusion}
\label{sec:conclusion}
In this paper, we consider the problem of maximizing an unknown costly-to-evaluate function when we have a small evaluation budget. Using the Bayesian optimization framework, we proposed a two-phase exploration-exploitation algorithm that finds the maximizer of the function with few function evaluations by leveraging the Lipschitz property of the unknown function. In the exploration phase, our algorithm tries to remove as many points as possible from the search space and hence shrinks the search space. In the exploitation phase, the algorithm tries to find the point that is closest to the optimal. Our empirical results show that our algorithm outperforms EI (even in its best condition).

\bibliographystyle{plainnat}
\bibliography{FiniteHorizon}

\appendix
\section{Proof of Lemma 1}

Let $f(x)$ be our function prediction at any point $x$ distributed as a normal random variable with mean $\mu_x$ and variance $\sigma^2_x$; i.e $f(x) \sim \mathcal{N} (\mu(x,\sigma^2_x))$ where $\mu_x$ and $\sigma^2_x$ obtained from Gaussian process. Suppose $y_{max}$is the best current observation, the probability of improvement of $I\in[0,M-y_{max}]$ can be calculated as $p(f(x)=y_{max}+I)$:

\begin{equation}
\label{eq:mei1}
p\Big(f(x)=y_{max}+I\Big)=\frac{1}{\sqrt{2\pi}\sigma_x} \text{exp}\left(-\frac{(y_{max}+I-\mu_x)^2}{2\sigma^2_x}\right). \\
\end{equation}  
Therefore we define $EI_M(x)$ as is simply the expectation of likelihood over $I\in[0,M]$ at any given point $x$: 

\begin{equation}
\label{eq:mei2}
\begin{aligned}
	EI_M(x)&= \int_{I=0}^{I=M-y_{max}} I\left\{\frac{1}{\sqrt{2\pi}\sigma_x} \text{exp}\left(-\frac{(y_{max}+I-\mu_x)^2}{2\sigma^2_x}\right)\right\}dI \\
	&=\frac{1}{\sqrt{2\pi}\sigma_x} \text{exp}\left(-\frac{(y_{max}-\mu_x)^2}{2\sigma^2_x}\right) \int_{0}^{M-y_{max}}I\, \text{exp}\left(-\frac{2I(y_{max}-\mu_x)+I^2}{2\sigma^2_x} \right) dI.
\end{aligned}
\end{equation}  
Let define

\begin{equation}
\label{eq:t}
\begin{aligned}
T&=\text{exp}\left(-\frac{2I(y_{max}-\mu_x)+I^2}{2\sigma^2_x}\right)\\
\frac{\partial T}{\partial I}&=-\frac{1}{\sigma^2_x}\left(IT+(y_{max}-\mu_xT)\right),
\end{aligned}
\end{equation}
therefore we can get

\begin{equation}
\label{eq:it}
IT=-(y_{max}-\mu_x)T -\frac{\partial T}{\partial I}\sigma^2_x.
\end{equation}

Using equations \ref{eq:it},\ref{eq:t},\ref{eq:mei2} we can get

\begin{equation}
\label{eq:mei3}
\begin{aligned}
EI_M(x)&=\frac{1}{\sqrt{2\pi}\sigma_x} \text{exp}\left(-\frac{\left(y_{max}-\mu_x\right)^2}{2\sigma^2_x}\right) \int_{0}^{M-y_{max}}IT\,\, dI\\
&=\sigma_x\phi\left(\frac{y_{max}-\mu_x}{\sigma_x}\right)\\
&-(y_{max}-\mu_x)\int_{0}^{M-y_{max}} \frac{1}{\sqrt{2\pi}\sigma_x} \text{exp}\left(-\frac{1}{2}\left(\frac{y_{max}+I-\mu_ x}{\sigma_x}\right)^2 \right) \,dI.
\end{aligned}
\end{equation}

Let 
\begin{equation}
I^*=\frac{y_{max}+I-\mu_x}{\sigma_x}, \qquad then \qquad dI^*=\frac{dI}{\sigma_x},
\end{equation}
then the equation \ref{eq:mei3} can be written as
\begin{equation}
\begin{aligned}
EI_M(x)&=\sigma_x\phi\left(\frac{y_{max}-\mu_x}{\sigma_x}\right)\\
&-(y_{max}-\mu_x)\int_{\frac{y_{max}-\mu_x}{\sigma_x}}^{\frac{ M-\mu_x}{\sigma_x}} \frac{1} {\sqrt{2\pi}} \text{exp}\left(-\frac{1}{2}I^{*2}\right) \,dI^* \\ 
&=\sigma_x\phi\left(\frac{y_{max}-\mu_x}{\sigma_x}\right)- \left[(y_{max}-\mu_x)\left(\Phi\left(\frac {M-\mu_x}{\sigma_x}\right)- \Phi\left(\frac{y_{max}-\mu_x}{\sigma_x}\right)\right)\right].
\end{aligned}
\end{equation} 

Let 
\begin{equation}
\nonumber
u_1=\frac{y_{max}-\mu_x}{\sigma_x}, \\
u_2=\frac{M-\mu_x}{\sigma_x},
\end{equation}

then we can finally drive the maximum expected improvement at any given point $x$ as
\begin{equation}
MEI(x)=\sigma_x\big(-u_1\Phi(u_2)+u_1\Phi(u_1)+\phi(u_1)\big),
\end{equation}

where $\Phi(\cdot)$ is the normal cumulative distribution function and $\phi(\cdot)$ is the standard nomal distribution.

\end{document}